\documentclass[12pt]{article}
\usepackage{newtxtext,newtxmath}

\usepackage{graphicx}
\usepackage{xcolor}
\usepackage{hyperref}
\usepackage[letterpaper,margin=1in]{geometry}

\usepackage[style=science,articletitle=true]{biblatex}
\renewenvironment{abstract}
	{\quotation}
	{\endquotation}

\date{}

\makeatletter
\renewcommand{\fnum@figure}{\textbf{Figure \thefigure}}
\renewcommand{\fnum@table}{\textbf{Table \thetable}}
\makeatother

\usepackage{url}

\newcommand{\modelname}{AncLoRA}

\def\papertitle{
	Directing large language models to follow the letter or spirit of the law
}

\title{\bfseries \boldmath \papertitle}

\author{
	Peng Qian$^{1\ast}$,
	Andrew Li$^{2}$,
	Sam Chen$^{2}$,\and 
    Sonia K. Murthy$^{2}$,
    Yonatan Belinkov$^{2,3}$,
    Tomer D. Ullman$^{1,2\ast}$
    \and
	\small$^{1}$Department of Psychology, Harvard University\and
	\small$^{2}$Kempner Institute for the Study of Natural and Artificial Intelligence, Harvard University
    \and
    \small$^{3}$Taub Faculty of Computer Science, Technion --- Israel Institute of Technology\\
	\small$^\ast$Corresponding authors. Email: pqian@fas.harvard.edu; tullman@fas.harvard.edu
}

\begin{document} 

\maketitle

\begin{abstract} \bfseries \boldmath

The distinction between the spirit and letter of the law is a central issue across research and everyday life, and a growing concern for building safe, intelligent machines. What is this distinction based on, and how can we develop machines that follow the intention behind a rule? We used targeted adaptation that made large language models prioritize the spirit or letter of the law. With minimal modifications, our method significantly changed LLM behavior across diverse measures, novel vignettes, real-world scenarios, and influential legal cases. An analysis of model internals revealed a low-dimensional space with three interpretable dimensions matching a formal pre-specified framework for the geometry of legal concepts. These findings show how legal thought in LLMs may be organized and directed.

\end{abstract}

\noindent

\begin{figure}[!t]
    \centering
    \includegraphics[width=\linewidth]{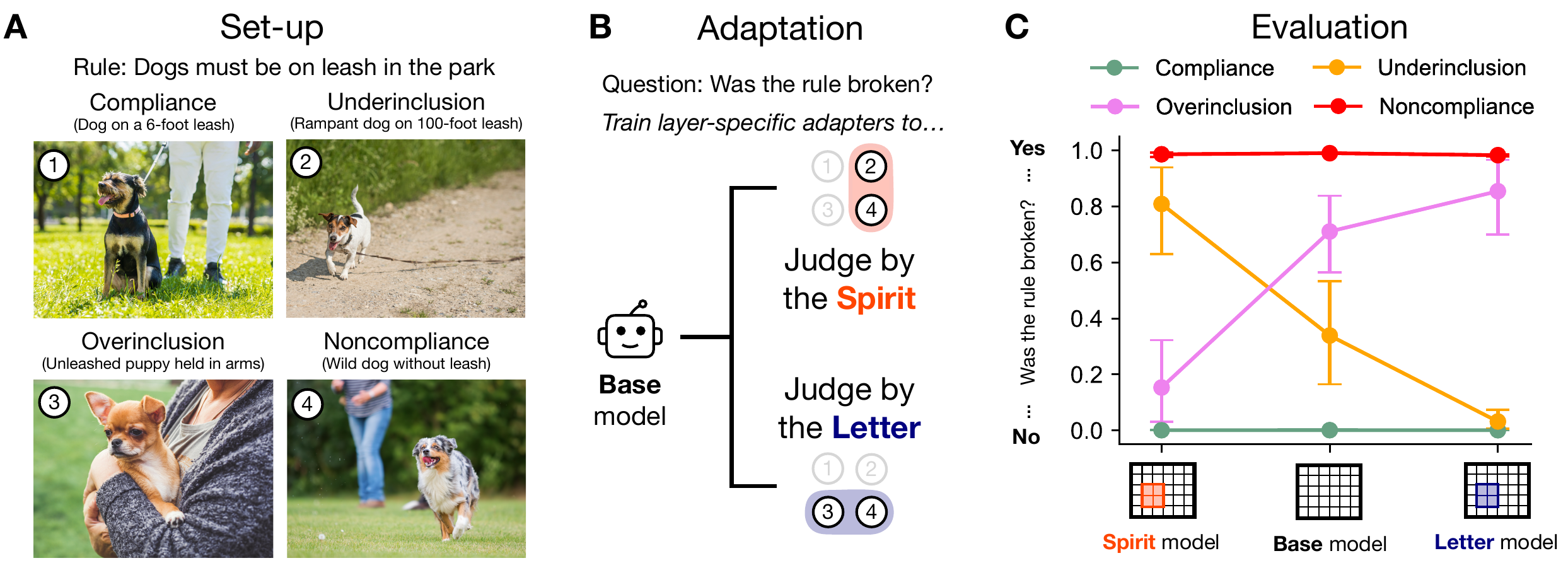}
    \caption{\textbf{Adapter training and evaluation pipeline.}  (\textbf{A}) Example vignette, rule, and behaviors. Responses broadly fit into four categories: Compliance (follow letter and spirit), Underinclusion (follow letter but not spirit), Overinclusion (follow letter but not spirit),  and (4) Noncompliance (violate letter and spirit). Our training included a set of such rules, vignettes, and behaviors.  (\textbf{B}) We trained layer-specific adapters with a main loss function of how closely the adapted model matched a spirit-based or letter-based judgment, with an added anchoring to base responses on factual understanding. (\textbf{C}) Applying the adapters systematically steered the base model to favor either a spirit-based or letter-based interpretation of the rules. Results are shown for held-out vignettes, with adapters applied at layer 32 of Qwen3.5-27B. Error bars represent 95\% Confidence Intervals.
        }
    \label{fig:intro-rule-rs}
\end{figure}

The distinction between the spirit of the law and the letter of the law is long-standing and far-reaching. For example, consider a simple rule like ``Dogs must be kept on leash in the park", alongside two cases: In the first case, a person carries a small unleashed puppy in their arms for their entire stay in the park. This person breaks the letter, but not the spirit of the law. In the second case, a person keeps their dog on a 100-foot leash, letting them run loose without control. This person technically complies with the letter, but breaks the spirit of the law (Fig. \ref{fig:intro-rule-rs}A). The distinction  is central to legal practice \cite{schulz1946history,scalia2012reading,barak2005purposive, hannikainen2022coordination}, has been of outstanding interest in philosophy \cite{aristotle2014nicomachean,hart2012concept} and religion \cite{aquinas2006summa28,hain2024circumventing}, and affects many everyday social interactions \cite{bridgers2025loopholes,bridgers2025learning}. This distinction is also a major and growing challenge for building machines that follow the intentions of their makers \cite{hadfield2019incomplete,kolt2026legal,liu2026large,o2025law}, and as such is of increasing concern for policymakers, legal professionals, engineers, and the public. Building more powerful and intelligent machines will not inherently cause them to follow either the spirit or letter of the law, just as an intelligent person may possess a full understanding of the intention of a rule and still find a way around it. How do large language models understand the distinction between letter and spirit, and can they be made to follow one or the other? 

We present an approach for directing large language models (LLMs) to prioritize either the letter or spirit of the law. Specifically, we propose an approach based on Low-Rank Adaptation with anchoring loss (\modelname), which induces targeted changes to model parameters. These changes push a model to prioritize either literal meaning or intended purpose when judging whether a behavior violated a rule. We assessed the effect of our method on a diverse set of stimuli, which included institutional and interpersonal scenarios, online postings about malicious compliance in everyday behavior, and real-world historical legal cases. Beyond the downstream effects on model behavior, we used our tools to analyze how basic legal concepts are constructed and represented in these models' internal states. Our analysis reveals a latent low-dimensional space with three separate, interpretable axes. The representational geometry of this space matched the predictions of a specific formal hypothesis for how the letter-and-spirit distinction should be organized. 

\subsection*{Learning lightweight adapters from minimal examples}

To illustrate the way in which we pushed language models to follow either the letter or spirit of the law, consider again the opening example about the dog-leash rule (Fig. \ref{fig:intro-rule-rs}A). Borrowing terminology used previously in legal studies \cite{tussman1949equal,schauer1991playing,struchiner2020experimental}, there are four basic types of behavior in response to a rule: (1) Compliance, where neither the letter nor spirit was broken, as when a person walks a dog on a standard 6-foot leash; (2) Underinclusion, in which the letter is technically followed but the spirit is subverted, such as a person who puts a dog on a 100-foot leash and lets them run around uninhibited; (3) Overinclusion, in which the letter of the law is broken but the intended spirit is kept, as in the case of a person carrying a small unleashed puppy in their arms; (4) Noncompliance, where both the letter and spirit are broken, as when someone lets an unleashed dog run amok. Under the strictest application of a letter-oriented approach, we would expect Overinclusion and Noncompliance to be grouped together as violations, while Compliance and Underinclusion would be grouped together as permissible. By contrast, a spirit-oriented approach to the law would group together Compliance and Overinclusion as permissible, and group Noncompliance and Underinclusion as violations. 

With these examples and distinctions in mind, we briefly describe our approach (technical details can be found in Materials and Methods in Supplementary Materials). We introduce a method based on Low-Rank Adaptation \cite{hu2022lora} that we term Anchored-LoRA (\modelname). This method uses a base model with frozen weights, and trains layer-specific adapters optimized to shift the model's responses. The main loss function for our \modelname{} quantified how close an adapted model's judgments came to those of an idealized letter-oriented or spirit-oriented approach. In addition to this main loss, the training objective consists of a soft anchoring loss. The anchoring loss penalizes an adapted model's deviation from the original responses of a base model for questions that relate to a basic understanding of the rule and the situation. This added constraint helps to train an adapter to intervene on targeted representations about legalistic ideas, without otherwise distorting the model's grasp of the scenario. The idea is that regardless of the legal approach one adopts, any judge should agree on basic questions such as whether the person who held the puppy in their arms put a leash on their dog.

Our training used a set of 57 hand-crafted vignettes about everyday rules. Each vignette crossed three stylistic variants of the task instructions, four types of behavior (Noncompliance, Overinclusion, Underinclusion, and Compliance), and three questions (a main question asking whether the behavior violated the rule, and two control questions to verify understanding). We used a 74\%/26\% train-evaluate split, with 1512 unique training prompts. For ease of computing the loss function, all prompts instructed a model to answer questions with a single-token response (Yes or No). However, as we show later in the results, the steering effect of the trained adapter generalized to free-form text responses and reasoning chains. Our main results report findings from a leading open-source LLM (at the time of writing), with native reasoning capability (Qwen3.5-27B). The Supplementary Materials detail similar results from another high-performance open-source model (Llama-3.1-70B-Instruct).

\subsection*{Evaluating letter/spirit adaptation across diverse domains}

We evaluated the generalization of our letter/spirit adaptation across diverse contexts. Unless stated otherwise, the statistical effects reported are based on Bayesian mixed-effect regressions. Further details regarding statistical analyses, evaluation prompts, example stimuli, and setup are in the Supplementary Materials. First, we tested the letter/spirit adapted models on the held-out rule vignettes (15 rules $\times$ 4 behavior types) with a novel prompt style not seen in the training process. As shown in Fig. \ref{fig:intro-rule-rs}C, with minimal changes to just one layer, adapter-active models systematically shifted their behaviors from the base model. When the spirit adapter was applied, Overinclusion was less likely to be judged as violating the rule ($\beta=-2.46$, 95\% CI: $[-3.25, -1.69]$), while Underinclusion was more likely to be judged as a violation ($\beta = 2.11$, 95\% CI: $[1.33, 2.94]$). When the letter adapter was applied, the adapted model was significantly less likely to judge an Underinclusion as a violation ($\beta=-1.24$, 95\% CI: $[-2.03, -0.45]$), and was more likely to judge cases of Overinclusion as similar to outright noncompliance ($\beta=0.58$, 95\% CI: $[-0.18, 1.37]$. This last difference was numerically positive but not statistically significant). 

Importantly, the adaptation towards letter- or spirit-based judgment did not come about from a general distortion in understanding. Both the spirit-adapted and the letter-adapted models maintained their judgments on Compliance ($\beta_\mathrm{letter} = -0.12$, 95\% CI: $[-0.84,0.61]$; $\beta_\mathrm{spirit} = -0.01$, 95\% CI: $[-0.76, 0.72]$) and Noncompliance ($\beta_\mathrm{letter} = -0.14$, 95\% CI: $[-0.90, 0.64]$; $\beta_\mathrm{spirit} = -0.02$, 95\% CI: $[-0.77,0.73]$ ). When prompted with control questions regarding the factual understanding of a scenario, the adapted models responded in a similar way to the base model (Supplementary Materials, Fig. \ref{fig:rule-question-comparison-qwen}). The adapted models also show no degradation on general commonsense knowledge, including professional knowledge of the law (Supplementary Materials, Table \ref{tab:mmlu-qwen35}). 

\begin{figure}[!t]
    \centering
    \includegraphics[width=\linewidth]{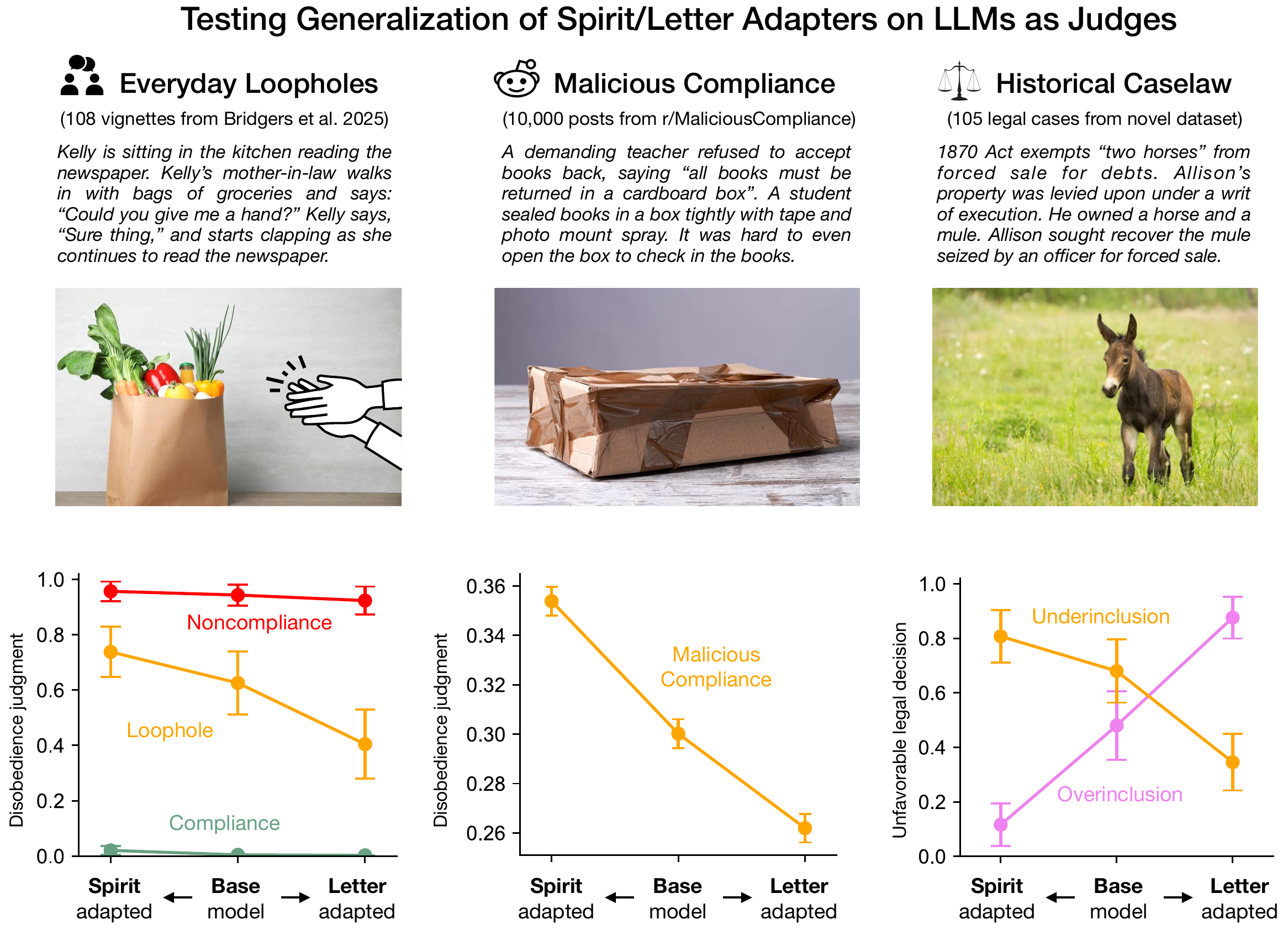}
    \caption{\textbf{Effect of letter/spirit adaptation across contexts and domains.} 
    All the results were based on base model Qwen3.5-27B and adapters trained to modify the Multi-layer Perceptron component of layer 32 of Qwen3.5-27B. 
    Top row shows snippets of stimuli from the domain of everyday loopholes (left), online posts from Reddit r/MaliciousCompliance (middle), and historical caselaw (right). Note that the original stimuli were rephrased or shortened here for illustrative purpose. Bottom row shows the effect of the spirit/letter adaptation for each domain. For the historical legal cases, models generated free-form responses, with additional inference-time amplification ($\eta=3.5$) of the adapter's influence. Error bars represent 95\% Confidence Intervals.
    } 
    \label{fig:rs-across-contexts}
\end{figure}

The results so far suggest that the steering induced by \modelname{} is both effective and targeted, when tested on hand-tailored held-out scenarios similar to those used in training. But how much does this adaptation generalize? We next evaluated spirit/letter adapted models on conceptually relevant, but out-of-distribution stimuli. We stress that no further training was carried out on the models beyond what was already described. Our first set of out-of-distribution stimuli drew on vignettes used in previous work to study intentional misunderstandings and the use of ``loopholes" to get around requests in everyday life \cite{bridgers2025loopholes,qian2024ambivalence}. These ``loopholes" stimuli move the focus from rules set by an authority to goals and requests in interpersonal relations. Similar to Underinclusion, loopholes exploit the ambiguity of language to subvert intended meaning. To see this, consider a mother who tells her child ``It's time to get in the bath.'' The child is playing with blocks, and does not want to stop. So, the child physically gets in the bathtub, without running the water, and continues to play. The request was followed in a technical sense, but not in the intended sense. The loophole stimuli set crosses 36 stories with three possible actions of the listener (compliance, loophole, noncompliance) and three power relationships (down, equal, up), for a total of 324 unique vignettes. We prompted our adapted models with all vignettes, and collected model judgments on whether the protagonist disobeyed the request. Aggregating over diverse social scenarios and power relations, we observe targeted steering effects similar to those seen for the original rule stimuli (Fig. \ref{fig:rs-across-contexts}A). The letter-adapted model was less likely to judge loophole behaviors as disobedience compared to the base model ($\beta=-1.0$, 95\% CI: $[-1.25, -0.75]$), and spirit-adapted model was more likely to judge them as disobedience ($\beta=0.51$, 95\% CI: $[0.25, 0.77]$). Again, judgments of Noncompliance and Compliance were largely unaffected by the application of adapters in either the letter direction ($\beta_\text{compliance} = -0.12$, 95\% CI: $[-0.40, 0.16]$; $\beta_\text{noncompliance} = -0.06$, 95\% CI: $[-0.35, 0.22]$) or spirit direction ($\beta_\text{compliance} = 0.15$, 95\% CI: $[-0.13, 0.43]$; $\beta_\text{noncompliance} = 0.13$, 95\% CI: $[-0.16, 0.41]$).

Our next set of out-of-distribution stimuli moved the assessment from hand-crafted vignettes to noisy, in-the-wild examples. Specifically, we curated a novel dataset of 10,000 online posts, sampled from the Reddit community \texttt{r/MaliciousCompliance}. This forum was created for people to share anecdotes of themselves or others ``conforming to the letter, but not the spirit, of a request''. The stories feature cases similar to loophole behaviors, as well as instances where people knowingly followed a wrong directive in order to annoy others. The examples varied considerably in linguistic style and narrative voice. Models were presented with an entire post, and judged whether a protagonist in the story disobeyed a directive. We emphasize that letter- and spirit-adapted models were not trained on these online posts, and any effect is driven by the original targeted training on the small set of curated examples. As shown in Fig. \ref{fig:rs-across-contexts}B, \modelname{} adapters steered the base model to shift its judgments, in a similar direction to that previously found for loopholes and Underinclusion ($\beta_\text{letter}=-0.32$, 95\% CI: $[-0.33, -0.30]$; $\beta_\text{spirit}=0.37$, 95\% CI: $[0.35, 0.38]$). The effect size is smaller than those found previously. This is likely due to the noisy nature of naturalistic data (see Supplementary Materials for more detail on noise in raw posts), but the overall changes are significant for both letter and spirit.

Up to this point, the results suggest that a targeted, minimal intervention could steer LLMs to be more spirit- or letter-oriented when judging everyday rules and requests. But does this effect extend to the formal legal system? To examine this, we developed a novel dataset of historical caselaw vignettes. We drew on various sources, including prominent cases cited by proponents of a textualist approach to jurisprudence \cite{scalia2012reading}, examples discussed by guidebooks in the common law tradition \cite{holland2013learning,jones2019introduction,sullivan2002sullivan}, cases reported in the news, and historical cases that involved disputes over statutory interpretation (based on source files from the Caselaw Access Project \cite{harvard2024caselaw}). Given the massive volume and variance of these historical cases, we translated them into comparable case summaries. In total, we crafted case summaries for 105 cases that covered diverse areas in law, of which 50 were Overinclusion and 55 were Underinclusion. We pre-registered this study and our hypothesized effects on Open Science Framework\footnote{Link to the pre-registration: \texttt{\url{https://osf.io/8gyvr/}}}. Unlike our previous evaluations in which models were instructed to give a single-token response, here we sampled free-form text responses. The pipeline is thus closest to a real judicial ruling, in which a judge may be inclined either towards the spirit or letter of the law. We again stress that the adapted models were not trained on any of these new stimuli or new patterns of response, having been adapted solely based on yes/no responses to a small set of hand-crafted vignettes. As shown in Fig. \ref{fig:rs-across-contexts}C, we observed a robust steering effect. As predicted in our pre-registration, letter-adapted model was much more likely to give unfavorable decisions to Overinclusion ($\beta=3.67$, 95\% CI: $[2.96,4.45]$), and favored Underinclusion ($\beta=-2.74$, 95\% CI: $[-3.33, -2.18]$). The pattern was flipped for spirit-adapted model: Overinclusion received more favorable rulings ($\beta=-3.62$, 95\% CI: $[-4.44,-2.87]$), while Underinclusion received more unfavorable rulings ($\beta=1.29$, 95\% CI: $[0.74,1.88]$). This effect persisted in explicit reasoning chains (Supplementary Materials, Fig. \ref{fig:qwen-thinking-caselaw-rs}).

\subsection*{A latent space of letter/spirit concepts in model internals}

Our results so far demonstrated that with a small training dataset, \modelname{} can induce systematic, robust, and specific control on model behavior that pushes models to follow either the letter or the spirit of the law. These steering effects were evaluated through model behavior. But what underlies this change? We examined model activation states following our use of \modelname{}, in order to study the latent representational geometry of letter-and-spirit in LLMs. As we detail below, we found a latent low-dimensional space with three interpretable axes, which matched a hypothesis for how the letter/spirit distinction should be organized. Our investigation of latent geometry was made possible by the use of open-source models that afforded fully transparent access to model internals. 

\begin{figure}[!t]
    \centering
    \includegraphics[width=\linewidth]{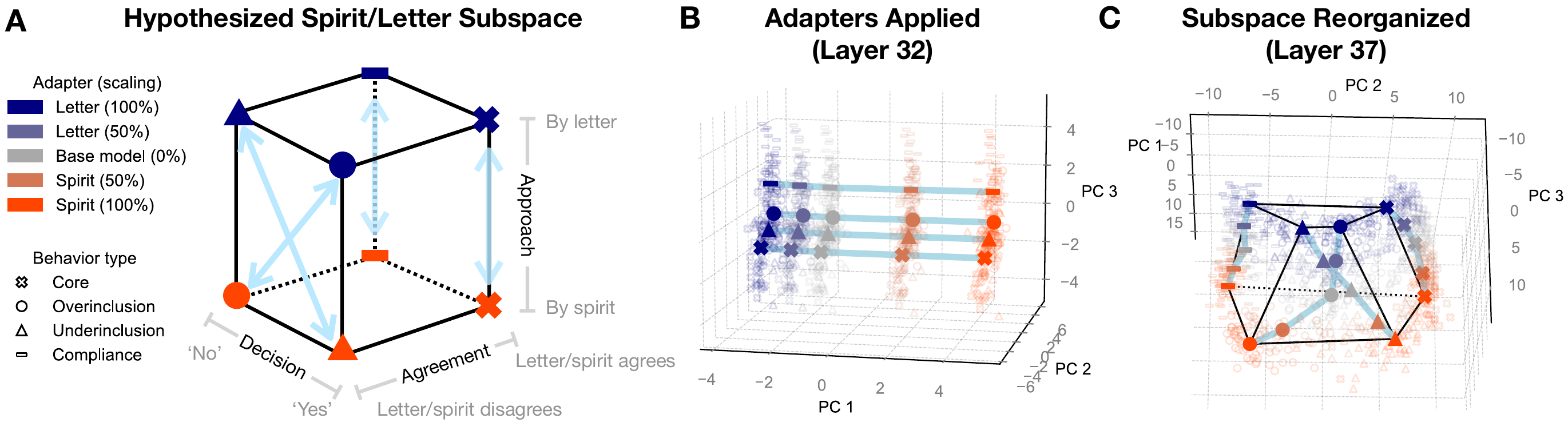}
    \caption{\textbf{Representational geometry of legal concepts in the model internals.} (\textbf{A}) A formal model of a conceptual space that may underlie the spirit/letter distinction in judgment. The space is defined by three separate dimensions: Decision (i.e. whether the behavior is judged as a violation or not), Agreement (whether the spirit and letter of the law conflict or not), and Approach (whether the judgment goes by the letter or spirit of the law). (\textbf{B}) Visualization of the low-dimensional space in the model residual streams across layers, after applying the trained adapters to Layer 32 of Qwen3.5-27B. Each hollow marker represents a specific item, while the larger solid marker represents the mean across the items. The light blue line connects the means across varying directions and scaling of the adapters. The first three principal components of the residual stream of the Transformer models reveal a structured low-dimensional manifold. The application of the adapters resulted in an immediate linear shift in the representational space. (\textbf{C}) After the subsequent nonlinear processing, later layers start to construct a space aligned with the formal theoretical model of the relevant legal concepts.} 
    \label{fig:representational-geometry-rs-panel}
\end{figure}

We first introduce our theoretical framing of a conceptual space that should underlie the letter/spirit divide. As is illustrated in Fig. \ref{fig:representational-geometry-rs-panel}A, a judgment of a particular behavior should be evaluated on three separate dimensions, which relate to three conceptual distinctions. These distinctions map onto key constructs in the philosophy of jurisprudence \cite{schauer1991playing}. The first dimension (``Decision'') tracks the final outcome, whether the behavior was classified as a violation or not. This captures the overt response. The second dimension (``Agreement'') tracks whether the letter and spirit of the law are in conflict. Along this dimension, Compliance and Noncompliance are at one end (spirit and letter are not in conflict), while Underinclusion and Overinclusion are on the other end (spirit and letter are in conflict). The third dimension (``Approach'') tracks whether the judgment is based on the letter or the spirit of the law. We note in particular that in this organization of the space, moving from one approach to another requires a non-linear shift, to correctly account for the change in Overinclusion and Underinclusion while maintaining the same judgment on Compliance and Noncompliance. We hypothesized that this conceptual representation would emerge as a consequence of our \modelname{} intervention, though other possible geometries exist (Fig. \ref{fig:rsa-comparison-qwen35}). We note that this particular structure would also account for why linear steering with contrastive pairs in this conceptual space would be ineffective in inducing a letter/spirit behavioral change in LLMs. 

To test our proposal regarding the representational space that may emerge in LLMs reasoning about the letter or spirit, we analyzed the internal activation vectors of Qwen3.5-27B with either the letter- or spirit- \modelname{} adapter applied at layer 32. We focused on this layer as the behavioral results showed that this was one of the layers in which the trained adapters induced robust and generalizable steering effects (see Supplementary Materials, Fig. \ref{fig:loophole-delta-effect-across-layers-qwen}). We weighted the adapters with different scaling coefficients, to either attenuate or amplify the effect of weight intervention (from the base model level of 0\%, to 50\%, to 100\%). We extracted the residual stream representations at the last token of the prompt, which is right before the model produces a response, given the vignettes in the original hand-crafted rules dataset. As a reminder, this prompt was presented in a new style unseen in the adapter training process. We then projected the activation vectors onto a low-dimensional space spanned by the first three principal components of the activation vectors. We chose to examine the first three components in accordance with our prior hypothesis, but note in addition that these components explained more than half of the total variance (for layer 37, PC1: 25.68\%, PC2: 19.71\%, PC3: 15.60\%).

As shown in Fig. \ref{fig:representational-geometry-rs-panel}, we found that after applying the adapters at the target layer, the LLM constructed a representational space that is qualitatively aligned with the dimensions of our framework for spirit/letter judgments. More specifically, the adapters resulted in a linear shift in the activation space (Fig.\ref{fig:representational-geometry-rs-panel}B), and after subsequent nonlinear processing, the residual stream output at layer 37 (and thereafter) started to form a low-dimensional manifold (Fig. \ref{fig:representational-geometry-rs-panel}C) that aligns with the dimensions of the theoretical framework (Fig. \ref{fig:representational-geometry-rs-panel}A). The observed representational structure suggests a global nonlinear transformation between letter-based judgment and spirit-based judgment. This nonlinear transformation is corroborated by the ineffectiveness of using contrastive pairs \cite{park2024linear,rimsky2024steering} to estimate a linear steering direction directly within this conceptual space, as well as the surprising effect of adapter subtraction (Supplemental Materials, Fig. \ref{fig:steering-comparison-held-out-rule-all} and \ref{fig:steering-comparison-loophole-all}). In addition to a qualitative comparison, we quantified the degree of representational alignment via representational similarity analysis (RSA, see Fig. \ref{fig:rsa-comparison-qwen35}). We measured the Spearman correlation between the representational similarity matrices of the empirical state space and that of various hypothetical models. The conceptual space we theorized achieved the highest Spearman correlation ($\rho$ = 0.76) compared to alternative hypotheses, and was significantly better than the next best model (for layer 37, $\Delta\rho=0.12$, 95\% bootstrapped CI: $[0.09, 0.15]$), suggesting that it better characterizes the latent low-dimensional structure inside the model's internal states.

\subsection*{Discussion}

We presented an approach for directing LLMs to prioritize either the spirit or letter of the law in judgment. By using a lightweight, targeted adaptation on the basis of a small number of vignettes, we induced a change in judgment that generalized across multiple domains and contexts. This included more spirit- or letter-based judgment in previously studied everyday social situations \cite{bridgers2025loopholes}, thousands of varied real-world reported examples of malicious compliance, and free-form judgment for influential caselaw. When analyzing the representational space that underlies changes in judgment, we found a low-dimensional space reorganized in accordance with our hypothesis about the possible structure of spirit/letter concepts. The low-dimensional space maps onto established concepts in legal theory \cite{schauer1991playing}, and its non-linear organization explains why alternative steering approaches based on direct contrast of exemplars would not induce a spirit--letter change in the LLMs.

Our work has several broad implications for multiple areas of research and engineering that study the understanding of rules and judgments, including AI and the law. A major current challenge in building intelligent machines that adhere to the intention of their makers is that such machines can find clever loopholes, workarounds, and in general follow the letter of the request rather than its intent \cite{lehman2020surprising,skalse2022defining,dharna2026ai,liu2026large}. The use of further rules and restrictions is unlikely to fully mitigate such behaviors, just as they have not mitigated human loopholes \cite{john2024dead}, which exist due to the intrinsic limits of natural language \cite{hart2012concept,hart1988incomplete, qian2024ambivalence} and complexity of social interactions \cite{ayotte2025loopholes}. Our work suggests a different path towards alignment, one that specifies a rule and then pushes a model towards possible approaches in interpretation. Such an approach can be used for positive alignment (telling a machine to follow a rule, with an added ``you know what I mean"), or malicious exploitation (giving a machine a rule, and pushing it to find all possible exploits).

Beyond building machines that better follow or subvert given guidelines, our work has important consequences for the deployment of AI models in caselaw, legal interpretation, and judgment. Given the success of LLMs, legal professionals have expressed both enthusiasm and concern regarding their use in legal interpretation \cite{arbel2024generative,grimmelmann2026generative}. Our work suggests that large language models can be made to exhibit opposing approaches to jurisprudence, with minimal perturbation. Our findings do not in themselves suggest one path forward for legal studies. Rather, it requires the incorporation of different normative legal standards and the further involvement of legal scholars in the development and steering of LLMs, while providing an interpretable approach for doing so. 

Our method showed that spirit-letter preferences can be induced and generalized, but this leaves open many questions for future work. Our own examples were varied, but they cast the models as judges evaluating the behavior of others. An important direction for future work should examine how preferences for the letter or spirit of the law (whether inherent or induced) affect how agents themselves take action. This is a complex issue that interacts with an agent's own incentive structure \cite{qian2024ambivalence}. It also highlights the need to understand the origins of the abuse of legal technicalities in AI models, such as whether they are learned or inherent, and whether they are driven by an actual motivation or a simple confusion. Beyond this, more work is needed to understand the inference of the spirit of a law (or rule, or request) from context. Our work shows models can be pushed in the direction of the spirit, but how one understands the intent or spirit without being told is a topic of much current interest \cite{levine2026resource}. Finally, our work leaves open the normative questions surrounding the right direction for spirit- or letter-based judgment, which lawmakers, judges, and the public may disagree on.

The tension between the letter and the spirit of the law is a major theme in human history, and a new challenge in the era of artificial intelligence. We hope that the current work can inform policy development, open new directions for AI alignment research, and lead to a better understanding of how the basic distinction between the letter and the spirit is organized in both minds and machines.

\clearpage

\printbibliography

\section*{Acknowledgments}

We thank members of the Computation, Cognition, and Development Labs at Harvard University as well as participants of the 52nd Annual Meeting of the Society for Philosophy and Psychology for their feedback on an earlier version of this work.

\paragraph*{Funding:}
This work was supported by the Kempner Institute for the Study of Natural and Artificial Intelligence. 

\paragraph*{Author contributions:}
Conceptualization: P.Q. and T.D.U.; Stimuli design: P.Q., A.L., S.C., S.K.M.; Investigation: P.Q., A.L., S.C.; Methodology: P.Q., S.K.M., Y.B., T.D.U.; Writing: P.Q., A.L., S.C., S.K.M., Y.B., T.D.U.

\paragraph*{Competing interests:}
There are no competing interests to declare.

\paragraph*{Data and materials availability:}
Stimuli, code, and experimental materials will be available at the following repository: \url{https://github.com/pqian11/letter-spirit-adapters}

\subsection*{Supplementary materials}
Materials and Methods\\
Supplementary Text\\
Figs. S1 to S12\\
Tables S1 to S2\\

\newpage

\renewcommand{\thefigure}{S\arabic{figure}}
\renewcommand{\theHfigure}{S\arabic{figure}}
\renewcommand{\thetable}{S\arabic{table}}
\renewcommand{\theequation}{S\arabic{equation}}
\renewcommand{\thepage}{S\arabic{page}}
\setcounter{figure}{0}
\setcounter{table}{0}
\setcounter{equation}{0}
\setcounter{page}{1}

\begin{center}
\section*{Supplementary Materials for\\ \papertitle}

    Peng Qian$^\ast$,
	Andrew Li,
	Sam Chen, 
    Sonia K. Murthy,
    Yonatan Belinkov,
    Tomer D. Ullman$^\ast$\\
	\small$^\ast$Corresponding authors. Email: pqian@fas.harvard.edu; tullman@fas.harvard.edu
\end{center}

\subsubsection*{This PDF file includes:}
Materials and Methods\\
Supplementary Text\\
Figures S1 to S12\\
Tables S1 to S2\\

\newpage

\section*{Materials and Methods}

\subsection*{Anchored Low-Rank Adaptation}

We trained layer-wise direction-specific Low-Rank Adapters (LoRA) with a multi-task learning objective. Formally, consider a base model $M$ (e.g. Qwen3.5-27B) with the weight parameters $\Theta_M$, the weights $\Theta_A$ of the adapter $A$ at the Layer $L$ of the model $M$, and the direction $d$ (letter or spirit) toward which the adapter $A$ would be trained. Consider a set of training examples that span the full combinations of scenarios $\mathcal{S}$, task instruction prompt styles $\mathcal{I}$, and the task question set $\mathcal{Q}$ (\textsc{violation}, \textsc{text}, \textsc{purpose}). We optimized the adapter parameters $\Theta_A$ with respect to the following multi-task loss function $\mathcal{L}_\mathrm{total}$:
\begin{align}
\mathcal{L}_\mathrm{total}=\frac{1}{|\mathcal{I}|\cdot|\mathcal{S}|}\sum_{i\in\mathcal{I}}\sum_{s\in\mathcal{S}}\ell_\mathrm{main}(\Theta_A)+\lambda\cdot\ell_\mathrm{anchor}(\Theta_A)
\end{align}
\noindent where the main task loss function $\ell_\mathrm{main}(\Theta_A)$ is defined as the cross-entropy loss between the target response token $w^\star$ and the adapted model's predicted probability distribution over the token vocabulary, given the model input $\mathcal{C}_{i,s,\text{\textsc{violation}}}$ consisting of scenario $s$, prompt style $i$, and the rule violation question $q=\text{\textsc{violation}}$. Since the probability mass of the target distribution is concentrated on one specific token $w^\star$ (which depends on the direction $d$, whether to go with the letter or spirit, as well as the type of behavior to be judged in scenario $s$), the cross-entropy loss can be simplified as the surprisal of the target token $w^\star$:
\begin{align}
&-\sum_{w\in\mathcal{W}}\mathbf{1}_{s,d,\textsc{violation}}[w=w^\star]\log P(w|\mathcal{C}_{i,s, \mathrm{\textsc{violation}}};\Theta_M,\Theta_{A}) \\
=&-\log P(w^\star|\mathcal{C}_{i,s, \mathrm{\textsc{violation}}};\Theta_M,\Theta_A) 
\end{align}
\noindent The anchoring loss $\ell_\mathrm{anchor}(\Theta_A)$ is defined as the sum of cross-entropy loss between the predicted token distribution of the base model $M$ and the predicted token distribution of the adapted model ($M+A$), for the same control questions $q\in \{\mathrm{\textsc{text}}, \mathrm{\textsc{purpose}}\}$ given a scenario $s$ and a task instruction prompt style $i$:
\begin{align}
\sum_{q\in \{\mathrm{\textsc{text}}, \mathrm{\textsc{purpose}}\} }-\sum_{w\in \mathcal{W}}\overbrace{P(w|\mathcal{C}_{i,s, q};\Theta_M)}^\text{base model as anchor}\log P(w|\mathcal{C}_{i,s, q};\Theta_M,\Theta_{A})
\end{align}
Conceptually, the behaviors of the base model $M$ on the control questions serve as the anchor, or in other words add a soft constraint to the space of possible adapter weight parameters $\Theta_A$. The anchoring loss term $\ell_\mathrm{anchor}(\Theta_A)$ penalizes $\Theta_A$ that heavily distorts the model's judgment on the control questions before and after adding the adapter $A$. Also note that the target of the anchoring loss term depends on the specific choice of base model $M$. We multiplied the anchoring loss $\ell_\mathrm{anchor}(\Theta_A)$ 
with a small coefficient $\lambda$ to balance the main loss and the anchoring loss. This multitask learning objective is designed for optimizing the adapter parameter $\Theta_A$ toward a targeted effect.

\subsubsection*{Adapter Training Setup}

Specifically, we added and trained layer-wise anchored LoRA adapter $A$ for the \texttt{up\_proj} matrix of the MLP module (Multi-Layer Perceptron) at Layer $L$ of a base model. We set the loss weighting parameter $\lambda$ as 0.1. We chose a rank of $16$ for the adapters based on an initial exploration of the hyperparameters, and set the adapter alpha coefficient as $32$ during training ($\alpha=2r$). We trained the adapter parameter $\Theta_A$ with the optimizer AdamW and a learning rate of $2\times 10^{-4}$ with a cosine schedule. Each adapter was trained for 3 epochs.

We selected \texttt{Llama-3.1-70B-Instruct} and \texttt{Qwen3.5-27B} as the base models for our experiments. Both models are representative examples of open-source Large Language Models, with good performance on major evaluation benchmarks comparable to their closed-source counterparts. \texttt{Qwen3.5-27B} also comes with native reasoning capacity and was built with a Mixture-of-Experts architecture. Most importantly, these open-source models support direct intervention on model parameters and mechanistic analysis of model internal representations. We accessed the base model weights via Hugging Face Transformers \cite{wolf2019huggingface}.

For both base models, we trained adapters for every other layer starting from the second to the last layer. That is, layer 2, 4, 6, ..., 80 for \texttt{Llama-3.1-70B-Instruct} and layer 2, 4, 6, ..., 64 for \texttt{Qwen3.5-27B}. Each layer-wise \modelname{} adapter was initialized and trained independently with respect to the objective function defined above. That is, we added a single adapter $A$ at layer $L$ for the base model, optimized and saved the adapter parameters, and removed it before training another adapter at a different layer $L'$. We trained separate sets of adapters for the letter direction and the spirit direction. The training examples consist of 504 vignettes, crossing $42$ scenarios, $4$ types of behaviors (compliance, underinclusion, overinclusion, and noncompliance), and $3$ different styles for formatting the task instruction prompts. The prompt used for training instructed the model to only output `\texttt{Yes}' or `\texttt{No}' token as the response to the question. The desired token depends on whether the direction is letter or spirit. We describe the Rule dataset and the training prompts in the following sections.

\subsection*{Rule Dataset}

We crafted 57 diverse scenarios about everyday rules. Among all the scenarios, 42 were used for adapter training, and the remaining 15 were held out for evaluation. Each scenario describes a specific rule, as well as the context in which it was set, and is paired with four types of behaviors: compliance (neither the letter nor the spirit was violated), underinclusion (spirit is violated, but not the letter), overinclusion (letter is violated, but not the spirit), and noncompliance (both letter and spirit are violated). In total, there are 228 unique stories (scenario--behavior combinations). For each scenario, we wrote three questions: a main question on rule violation judgment (\textsc{violation}), and two control questions on judgment of aspects of the behavior related to the rule's text (\textsc{text}) or purpose (\textsc{purpose}). Each vignette starts with the background to introduce the rule, continues with one of the four types of behaviors, and ends with one of the three questions. Below is an example scenario from the stimuli: 

\begin{quote}
    The community park at Sunnyville has a playground, a skatepark, a baseball field, and a large grass area. People enjoy the park very much and some bring their dogs as well. To prevent unexpected dog fights as well as chaos caused by running dogs, the neighborhood association announces a rule: ``Dogs must be kept on leash in the park''.\\

    \texttt{\textsc{[noncompliance]}}
    Natalia brings her dog to the community park. Natalia's dog is not on leash, runs around, and chases people in the skatepark.

    \centerline{\texttt{\textsc{or}}}

    \texttt{\textsc{[overinclusion]}}
    Natalia brings a puppy to the community park. The puppy is very small and not on leash. Natalia holds the puppy in her arms for the whole time.

    \centerline{\texttt{\textsc{or}}}

    \texttt{\textsc{[underinclusion]}}
    Natalia brings her dog to the community park. Natalia keeps the dog on an extremely long 200-foot leash. Without effective control, the dog runs wildly and chases skaters in the skatepark.

    \centerline{\texttt{\textsc{or}}}
    
    \texttt{\textsc{[compliance]}}
    Natalia brings her dog to the community park. The dog is kept on a typical 6-foot leash. Natalia makes sure that the dog behaves and walks on the paved path.\\

Question (rule violation): Did Natalia break the rule?

Question (rule's text): Did Natalia fail to keep the dog on a leash?

Question (rule's purpose): Did Natalia fail to control the dog's running to avoid impact on others?
\end{quote}

For the main task question (\textsc{violation}), the target response is the same across judgment approaches for cases of noncompliance and compliance. Regardless of the direction of the adapter, the model should respond `\texttt{Yes}' for noncompliance cases and `\texttt{No}' for compliance cases. But for cases of underinclusion, we expect a model to more likely respond `\texttt{No}' when steered toward following the letter of the law, and `\texttt{Yes}' when steered toward following the spirit of the law. For cases of overinclusion, we expect the opposite pattern: a model should more likely respond `\texttt{Yes}' when steered toward following the letter of the law, and `\texttt{No}' when steered toward following the spirit of the law.

For the control questions (\textsc{text} and \textsc{purpose}), we expect models to give similar answers regardless of the direction of the learned adapters. As a concrete example, for the case of underinclusion given the dog leash rule described above, we expect models to generally respond `\texttt{No}' when asked about whether the person failed to keep the dog on a leash, since the dog was in fact on a leash.

\subsection*{Loophole Dataset}

We drew on experimental materials developed by prior studies on intentional misunderstandings \cite{bridgers2025loopholes}. These materials were originally developed for behavioral studies with people. There are 36 sets of scenarios. Each scenario describes a social interaction between two characters, where the main character (the listener or actor) responds to the directive or request that was given by their social partner (the speaker). Each scenario is paired with three different responses from the listener: compliance, outright disobedience, or a loophole. Each scenario also has three variants where the two characters are in different power relationships, \textsc{up} (the speaker has more power than the listener), \textsc{equal} (the speaker and the listener have equal power), or \textsc{down} (the speaker has less power than the listener). A specific vignette starts with the social context and the speaker's request, continues with one of the three kinds of actions from the listener, and ends with the task question on whether the listener disobeyed the request. We present an example item as follows:

\begin{quote}
    Chen is working on his laptop in the living room.
    
    Chen's father comes into the room and tells him:

    ``It's time to take the dog for a walk.''

    Chen's father leaves the room.\\

    \texttt{\textsc{[noncompliance]}}
    Chen does not take the dog for a walk.

    \centerline{\texttt{\textsc{or}}}

    \texttt{\textsc{[loophole]}}
    Chen takes the dog for a 1-minute walk to the end of the driveway and back.

    \centerline{\texttt{\textsc{or}}}

    \texttt{\textsc{[compliance]}}
    Chen takes the dog for a 15-minute walk around the neighborhood.\\

    Chen's father finds out what Chen did.\\

    \texttt{\textsc{question}}\\
    Did Chen disobey the request of Chen's father?
\end{quote}

In the example above, Chen was given a request by Chen's father (up power relationship). In other variants, Chen was given the same request by Chen's boyfriend (equal power relationship) or Chen's son (down power relationship).

The original survey materials are stored in the Open Science Framework repository of the human behavioral studies of loopholes \cite{bridgers2025loopholes} (\texttt{\url{https://osf.io/rwgmx}}).

\subsection*{Reddit r/MaliciousCompliance Dataset}

We collected a novel dataset of Reddit posts from the subreddit \texttt{r/MaliciousCompliance}\footnote{\texttt{\url{https://www.reddit.com/r/MaliciousCompliance/}}}. The r/MaliciousCompliance subreddit is an online community featuring shared stories of ``people conforming to the letter, but not the spirit, of a request''. We used the Arctic Shift API\footnote{\texttt{\url{https://github.com/ArthurHeitmann/arctic_shift}}} to find posts from Reddit automatically. We submitted multiple API requests to fetch posts that occurred from the beginning of the subreddit up until December 2025 and removed global duplicates. We also applied a constraint of a minimum of 20 upvotes to ensure that the collected posts had generated some interest in the online community. In total, we retrieved 10000 posts.

From a random sample of the Reddit posts, we noticed a high variance in the content quality of the posts. Some of the shared stories have a general flavor of malicious intent, but did not clearly fit into the category of ``conforming to the letter but not the spirit''. For example, below is a post\footnote{ \texttt{\url{https://www.reddit.com/r/MaliciousCompliance/comments/1ol8u8z/use_slow_computer_for_demanding_project/}}} that is unclear in terms of fitting the category of malicious compliance despite thousands of upvotes:

\begin{quote}
I got voluntold for the job of switching a paper-based corporate learning to computer-based, including web based training. I did not have a desk or a computer, so I brought in my personal laptop. The boss objected and stated I needed to write a business case for a computer.

A week later I got the absolute minimum system that met the minimum requirements on the box. I started the painful process of converting a Powerpoint into an Adobe Captivate file. When it came time to compile the first file, the computer stated it would be three hours before it finished, maybe, so I headed to the breakroom.

The executive director for the project happened to walk in and asked me what I was doing there. “I’m staying not frustrated while waiting for the first draft to compile, should be about another two hours sir.” It was five hours.

When I showed up the next day, my computer had been upgraded to the then top model with dual monitors.

The next day, my fancy unit was on the boss’ desk, and I had his even older, slower computer. This time compiling was over ten hours. Back to the breakroom. Same executive walks in, I just smile, nod, and go back to my lunch.

The next day, I had two computers on my desk, the still compiling boss’ unit and my previously issued fancy one. The boss was cleaning out his desk having been sent back to frontline, non-boss work.

It felt so good to give that company the boot once the project completed.
\end{quote}

\noindent Nevertheless, we believe this is a useful case for assessing the effect of the trained adapters on noisy data in the wild. 
For each Reddit post, the Large Language Models were given a question in generic phrasing: ``Did the protagonist disobey the request they were given?''

\subsection*{Historical Legal Case Dataset}

We developed a dataset of historical legal cases for evaluating model generalization. The dataset consists of summaries of 105 real-world legal cases drawn from prominent examples used in establishing relevant legal doctrine \cite{scalia2012reading}, cases in the United States as sourced from the Caselaw Access Project \cite{harvard2024caselaw}, as well as cases from the broader common law legal systems. The Caselaw Access Project data was sourced from the Common Pile dataset \cite{kandpal2025common}, hosted on Hugging Face (\texttt{\url{https://huggingface.co/datasets/common-pile/caselaw_access_project}}). These cases cover a diverse range of areas in law, including civil, business, and public issues. Among the 105 cases, 50 cases are categorized as overinclusion (breaking the letter but not the spirit), and the other 55 cases are categorized as underinclusion (breaking the spirit but not the letter).

We note that the category assigned to a case depends on the phrasing of the judgment question and the specific party involved in the dispute. For example, in one case (United States v. Perry, 133 F. 841) the central dispute is about whether a specific artwork falls within the scope of the statutory language. The factual information shows that the artwork does not come within the letter of the law, but is likely covered under the legislative intent. From the perspective of the importer, claiming the tax benefit for this artwork would be considered ``underinclusive''. The claim does not technically satisfy the statutory definition, so the importer would \textit{not} be entitled under the strict letter of the law, but would be entitled under its spirit. Therefore, given a question regarding whether the importer was entitled to the exemption, going with the spirit of the law would mean answering ``Yes'' and going with the letter would mean answering ``No''.  If we were to phrase the question as about whether the importer is ineligible for the reduced tax rate, then we would flip the answers of ``Yes'' and ``No'', and thus classify the case as overinclusion instead. The summary and question for this case are shown as follows:

\begin{quote}
The provision in paragraph 454, Tariff Act July 24, 1897, c.11, section 1, Schedule N, 30, Stat. 151 specifies duty rate for artworks, specifically “paintings in oil or water colors, pastels, pen and ink drawings, and statuary”. The Act defines the term “statuary” as including “only such statuary as is cut, carved, or otherwise wrought by hand from a solid block or mass of marble, stone, or alabaster, or from metal, and as is the professional production of a statuary or sculptor only”. Articles falling under this definition would be subject to a reduced duty rate of 15 percent ad valorem, pursuant to a reciprocal commercial agreement with Italy. General manufactures of marbles would be subject to a higher duty rate at 50 percent ad valorem, under paragraph 115 of the Act, Schedule B. The specific definition of “statuary” in paragraph 454 of the Act intended to distinguish carved sculptures of solid marble from those statuary made from composition stone or similar materials by molding.

Perry, Ryer \& Co. and Wirt Leland imported two antique marble statues into the United States from Italy. The statues are about 150 to 200 years old, and represent a faun and Flora. The net values of the two statues are 3000 francs. Both statues were produced by a professional sculptor in Italy. Each statue is about eight feet in height. The statues are each of three pieces, and were each sculptured from three separate blocks of marbles for convenience in handling, instead of carved out of a single block of marble. It is typical for many of the best statues of a heroic size to be executed from more than one piece of marbles.

The two marble statues were assessed at a duty rate of 50 per cent ad valorem by the Collector of Customs at the port of New York. Perry, Ryer \& Co. and Wirt Leland claimed these two articles to be dutiable at 15 per cent. The duty assessment was brought into court.

\texttt{\textsc{question}}\\
If you are the judge to rule this case, do you think the two Italian statues imported by Perry, Ryer \& Co. and Wirt Leland are entitled to the reduced duty rate under the provision in paragraph 454, Tariff Act July 24, 1897?

\end{quote}

All the stimuli were handcrafted and fact-checked by the authors of the paper. Some of the summaries may include sentences or phrases that were directly imported from the original source of the court decisions for precise statement of the facts. We used brief case summaries generated by off-the-shelf Qwen3.5-27B over the original court decision/opinion texts to facilitate the process of selecting relevant cases from the Caselaw Access Project, but in the final vignettes we composed a new summary based on information in the original legal documents. We also searched for the historical versions of the relevant statutes through library access and digitized legal databases as an additional verification of the case information.

\subsection*{Prompts for Constructing Training Set}

We designed three stylistic variations of the task instruction prompts for training the Anchored LoRA adapters. Here we show the system prompt and user prompt separately for clarity of presentation. These prompts were combined together into the chat template format via the corresponding tokenizer of the Large Language Models we examined here. In the final complete vignette, the placeholder \texttt{\{context\}} is replaced by the paragraph that sets the background for the rule, \texttt{\{behavior\}} is replaced by the description of a kind of behavior (compliance, underinclusion, overinclusion, or noncompliance), and \texttt{\{question\}} is replaced by one of the three questions (\textsc{violation}, \textsc{text}, or \textsc{purpose}). The Rule Dataset section shows an example stimulus.

\noindent Prompt variant \textbf{A}:
\begin{quote}
\small
    \texttt{\textsc{system prompt}}\\
    Please carefully read the scenario and answer the following Yes or No question about the person's behavior following the scenario. Answer only with ``Yes" or ``No".
    
    \texttt{\textsc{user prompt}}\\
    Scenario: \texttt{\{context\}} Behavior: \texttt{\{behavior\}} Question: \texttt{\{question\}}
\end{quote}

\noindent Prompt variant \textbf{B}:

\begin{quote}
\small
    \texttt{\textsc{system prompt}}\\
    Please carefully read the scenario and answer the following Yes or No question about the person's behavior following the scenario. Answer only with ``Yes" or ``No".
    
    \texttt{\textsc{user prompt}}\\
    The scenario is as follows. \texttt{\{context\}} Later on, the following happens. \texttt{\{behavior\}} \texttt{\{question\}}
\end{quote}

\noindent Prompt variant \textbf{C}:

\begin{quote}
\small
    \texttt{\textsc{system prompt}}\\
    Please carefully read the scenario and answer the following Yes or No question about the person's behavior following the scenario. Answer only with ``Yes" or ``No".
    
    \texttt{\textsc{user prompt}}\\
    Background information:\texttt{\textbackslash n\textbackslash n\{context\}\textbackslash n\textbackslash n}Incident report:\texttt{\textbackslash n\textbackslash n\{behavior\}\textbackslash n\textbackslash n} Decision question (answer Yes/No):\texttt{\textbackslash n\textbackslash n\{question\}}
\end{quote}

\subsection*{Prompts for Constructing Test Set}

Here we show the prompts for model evaluation. We show the system prompt and user prompt separately for clarity of presentation. These prompts were combined together into the chat template format via the corresponding model tokenizer. 

\subsubsection*{Rule}

We use the following prompt for the rule violation judgment question:

\begin{quote}
\small
    \texttt{\textsc{system prompt}}\\
    You are a helpful AI assistant.
    
    \texttt{\textsc{user prompt}}\\
    Please read the following scenario and judge whether a person's behavior violated the rule. Please only answer `Yes' or `No'.\texttt{\textbackslash n\textbackslash n\{context\}\textbackslash n\textbackslash n\{behavior\}\textbackslash n\textbackslash n\{question\}}
\end{quote}

We use the following prompt for the control questions regarding the rule's text or purpose:

\begin{quote}
\small
    \texttt{\textsc{system prompt}}\\
    You are a helpful AI assistant.
    
    \texttt{\textsc{user prompt}}\\
    Please read the following scenario and answer the question. Please only answer `Yes' or `No'.\texttt{\textbackslash n\textbackslash n\{context\}\textbackslash n\textbackslash n\{behavior\}\textbackslash n\textbackslash n\{question\}}
\end{quote}

\subsubsection*{Loophole}

\begin{quote}
\small
    \texttt{\textsc{system prompt}}\\
    You are a helpful AI assistant.

    \texttt{\textsc{user prompt}}\\
    Please read the scenario below and judge whether the protagonist's behavior (\texttt{\{actor\}}) disobeys the request of the other person (\texttt{\{actor\}}'s \texttt{\{other\_person\}}) in the scenario. Please only respond with `Yes' or `No'.\textbackslash n\textbackslash n\texttt{\{story\}}\textbackslash n\textbackslash n\texttt{\{question\}}
\end{quote}

\subsubsection*{Reddit r/MaliciousCompliance}

\begin{quote}
\small
    \texttt{\textsc{system prompt}}\\
    You are a helpful AI assistant.

    \texttt{\textsc{user prompt}}\\
    Please read the following Reddit post, and evaluate whether the protagonist's response disobey the request they were given. Please only output `Yes' or `No' to indicate your answer.\texttt{\textbackslash n\textbackslash n}Reddit post:\texttt{\textbackslash n\{post\}\textbackslash n}Did the protagonist disobey the request they were given? Please only output `Yes' or `No' to indicate your answer.
\end{quote}

\subsubsection*{Historical legal cases}

The prompt used for eliciting model responses on historical legal cases is as follows:

\begin{quote}
\small
\texttt{\textsc{system prompt}}\\
You are a helpful AI assistant.

\texttt{\textsc{user prompt}}\\
Please read the following case and answer the question.\texttt{\textbackslash n\textbackslash n\{case\_summary\}\textbackslash n\textbackslash n\{question\}}

\end{quote}

We used the off-the-shelf Qwen3.5-27B, a leading open-source Large Language Model, to automatically annotate the category of the generated free-form responses from both the base and adapted models. The prompt used for annotating free-form responses is as follows:

\begin{quote}
\small
\texttt{\textsc{system prompt}}\\
You are a helpful AI assistant.

\texttt{\textsc{user prompt}}\\
You will be reading some paragraphs below that indicate a response to a question. The question that the following paragraphs respond to is: ``\{\texttt{question}\}".\texttt{\textbackslash n\textbackslash n}Your task is to just focus on the content of the following paragraphs and extract the answer as articulated in the paragraphs below, i.e. whether in conclusion the response indicates `Yes' as the answer to the question or `No' as the answer to the question. Your task is simply to extract the answer. Please just output `Yes' or `No' based on your assessment of the response as indicated in the paragraphs below. Your task is to simply extract the answer written in the paragraphs below and please do not confuse it with your own judgment. The paragraphs to be processed are as follows:\texttt{\textbackslash n\textbackslash n \{generated\_response\}}
\end{quote}

\subsection*{Model Evaluation Setup}

For evaluation prompts that instruct a Large Language Model to respond just with a single token `Yes' or `No', we directly extracted the logits for these two tokens from the model's last layer and computed the normalized probability distribution over `Yes' and `No' as the next predicted token.

For free-form text generation, we sampled 5 random generation from the models, with the maximum number of new tokens set to 1000. For Qwen3.5-27B with native thinking capacity enabled, we set the maximum number of tokens used for the reasoning chain to 4000. We used the default sampling parameters (temperature, \texttt{top\_k}, \texttt{top\_p}) set by the config file of each model respectively. The generated responses from both base and adapted models were then annotated by the same off-the-shelf model Qwen3.5-27B, to map the free-form text to a categorical response of `Yes' or `No'. The annotator model (off-the-shelf Qwen3.5-27B, with no further fine-tuning) received the generated texts from other models as its input, together with an instruction prompt, and was instructed to extract the stated answer as `Yes' or `No'. We then computed the normalized probability distribution over the `Yes' and `No' token, and mapped it to `Yes' if $P(\text{`Yes'}|\text{generated\_response})>0.5$, and `No' otherwise. Among a sample of 100 free-form responses across various cases, we manually checked the annotations, and found that this pipeline accurately extracted all the answers.

\subsection*{Statistical Analyses}

We ran Bayesian mixed-effect ordered beta regressions to analyze the letter/spirit adapter effects, for cases where the model directly output a normalized probability distribution over the token `\texttt{Yes}' and `\texttt{No}'. These analyses applied to model responses on the Rule dataset, Loophole dataset, and the r/MaliciousCompliance dataset. The analyses were implemented through \texttt{R} packages \texttt{ordbetareg} \cite{kubinec2023ordered} and \texttt{brms} \cite{burkner2017brms}.

For \textbf{Rule} stimuli, we used the following model formula: 
\begin{center}
    \texttt{yes\_prob $\sim$ direction*behavior\_type + (1 | group)}
\end{center}
where the dependent variable is the normalized output token probability for `\texttt{Yes}' given the prompt. The categorical predictor `direction' for the adapter has three levels: base (no adapter was added), letter (Letter adapter), and spirit (Spirit adapter). We set `base' as the reference level. The categorical predictor `behavior\_type' codes the type of behavior, with four levels: compliance, noncompliance, overinclusion, and underinclusion. We set `compliance' as the reference level. We included a random effect of `group', which groups scenarios of specific behaviors based on the rule context. 

For \textbf{Loophole} stimuli, we used the following model formula: 
\begin{center}
    \texttt{yes\_prob $\sim$ direction*behavior\_type + power + (1 | story)}
\end{center}
where the dependent variable is the normalized output token probability for `\texttt{Yes}' given the prompt. The categorical predictor `direction' for the adapter has three levels: base (no adapter was added), letter (Letter adapter), and spirit (Spirit adapter). The level `base' was set as the reference level.  The categorical predictor `behavior\_type' codes the type of behavior, with three levels: compliance, loophole, and noncompliance, with compliance as the reference level. We included a fixed-effect predictor `power', as prior behavioral studies with human subjects suggested that people are sensitive to the power relationship when evaluating loophole behaviors \cite{bridgers2025loopholes,qian2024ambivalence}. The 36 different stories were included as a random effect.

For \textbf{Reddit r/MaliciousCompliance} stimuli, we used the following model formula:
\begin{center}
    \texttt{yes\_prob $\sim$ direction + (1 || post)}
\end{center}
where the dependent variable is the normalized output token probability for `\texttt{Yes}' given the prompt. The categorical predictor `direction' for the adapter has three levels: base (no LoRA was added), letter (Letter adapter), and spirit (Spirit adapter). The level `base' was set as the reference level. The post ID was included as a random effect.

For \textbf{Historical Legal Case} stimuli, we elicited free-form text responses from models and further annotated these sampled text as a binary response (`Yes' or `No') via an LLM-as-evaluator approach. As a result, we analyzed model responses with Bayesian mixed-effect logistic regressions. The model formula is as follows:
\begin{center}
    \texttt{response $\sim$ direction*behavior\_type + (1 | case\_id)}
\end{center}
The dependent variable `response' is a binary variable, where `Yes' (i.e. judgment or rulings against a specific party) and `No' (i.e. judgment or rulings in favor of a specific party) were coded as 1 and 0 respectively. We included two fixed-effect predictors and their interactions. The predictor `direction' for the adapter has three levels: base (no adapter was added), letter (Letter adapter), and spirit (Spirit adapter), with `base' set as the reference level.  The predictor `behavior\_type' has two levels: overinclusion and underinclusion, with `overinclusion' as the reference level. The legal case ID was included as a random effect.

In general, we ran model fitting with four MCMC chains, 4000 warm-up iterations and 4000 samples. Across the paper, we reported the estimated posterior median of the relevant coefficients, as well as their 95\% Credible Intervals. To decide significant effects, we examined whether the 95\% Credible Interval of a parameter includes 0.

\section*{Supplementary Text}

\subsection*{Additional Results of Letter/Spirit Adaptation}

\subsubsection*{Systematic Generalizations with \texttt{Llama-3.1-70B-Instruct}}

\begin{figure}
    \centering
    \includegraphics[width=\linewidth]{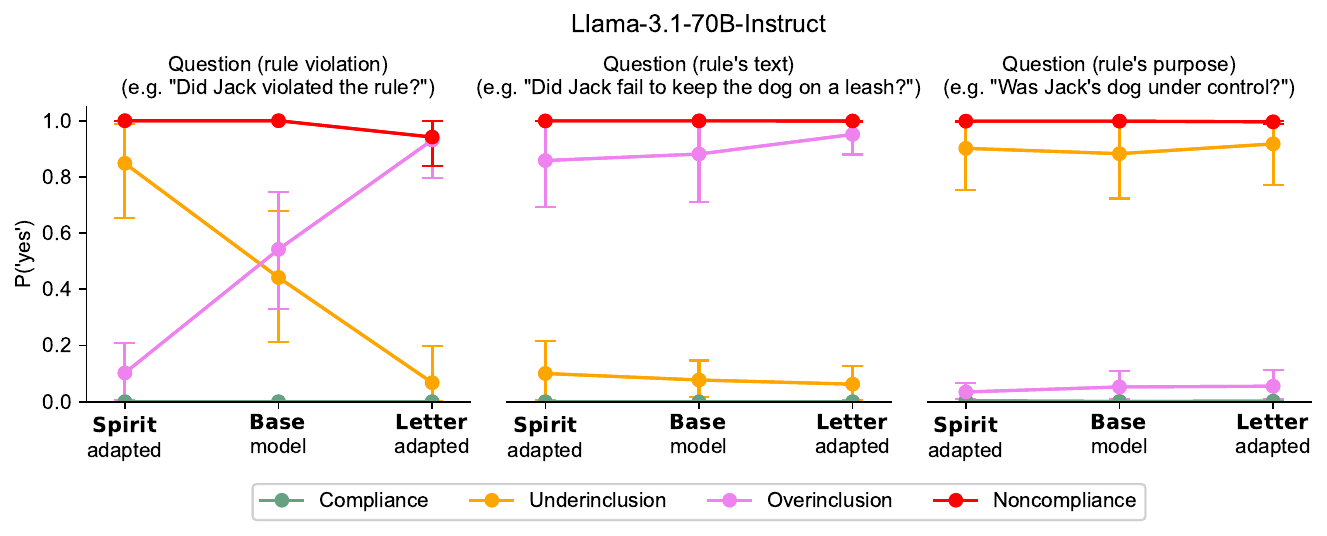}
    \caption{\textbf{Comparing adapted models across different types of questions on held-out data for \texttt{Llama-3.1-70B-Instruct}.} Base models and adapted models were evaluated on the main question (rule violation, left panel) and two control questions (rule's text and rule's purpose, middle and right panel). Letter- and spirit-oriented adapters induced a targeted steering effect on model's rule violation judgment, but not on control questions about the text and purpose of the rules. The $y$-axis shows the normalized probability of a model predicting the token `Yes' given the corresponding question. Error bars represent 95\% Confidence Intervals.}
    \label{fig:rule-question-comparison-llama}
\end{figure}

We evaluated layer-specific letter and spirit adapters for Llama-3.1-70B-Instruct across diverse contexts and domains. On held-out rule vignettes, the adapters induced a systematic effect similar to that for Qwen3.5-27B. As shown in Fig. \ref{fig:rule-question-comparison-llama}, compared to the base model, the letter-oriented adapter (applied at Layer 26) significantly increased the likelihood of rule violation judgment for overinclusion ($\beta = 1.89$, 95\% CI: $[0.814,3.047]$) and decreased the likelihood of rule violation judgment for underinclusion ($\beta = -1.27$, 95\% CI: $[-2.13, -0.44]$). We observed no steering effect on rule violation judgment for compliance ($\beta_\text{letter} = 0.026$, 95\% CI: $[-0.73, 0.76]$; $\beta_\text{spirit} = -0.064$, 95\% CI: $[-0.80, 0.67]$). For noncompliance, there was a significant decrease in judgment in the letter direction ($\beta = -3.65$, 95\% CI: $[-8.12,-0.96]$), although the change on the response scale is small. In the spirit direction, there was no significant effect ($\beta_\text{spirit} = 2.61$, 95\% CI: $[-3.04, 10.92]$).

We also examined the effect of letter/spirit \modelname{} adapters on the control questions regarding the rule's text or purpose. As shown in Fig. \ref{fig:rule-question-comparison-llama}, the adapter-active model responded to these questions in a similar way as the base Llama-3.1-70B-Instruct. For questions concerning the rule's text, such as ``Did Jack fail to keep the dog on a leash'' given the rule ``Dogs must be kept on leash in the park'', we found no significant differences between the adapted models and the base model for all the behavior types in both the letter direction (compliance, $\beta = 0.11$, 95\% CI: $[-0.60, 0.84]$; noncompliance, $\beta = 0.07$, 95\% CI: $[-2.07, 2.33]$; overinclusion, $\beta =  1.05$, 95\% CI: $[-0.15, 2.22]$; underinclusion, $\beta = 0.003$, 95\% CI: $[-0.75, 0.75]$) and the spirit direction (compliance, $\beta = 0.06$, 95\% CI: $[-0.66,0.75]$; noncompliance, $\beta=0.11$, 95\% CI: $[-2.01, 2.37]$; overinclusion, $\beta=-0.16$, 95\% CI: $[-1.21,    0.82]$; underinclusion, $\beta=0.06$, 95\% CI: $[-0.68, 0.81]$). For questions regarding a rule's purpose, such as ``Was Jack's dog under control?'' given the rule ``Dogs must be kept on leash in the park'', we also found no significant differences across the adapted models and the base model for all the behavior types in both the letter direction (compliance, $\beta=0.06$, 95\% CI: $[-0.66, 0.77]$; noncompliance, $\beta = -0.04$, 95\% CI: $[-1.45, 1.34]$; overinclusion, $\beta = 0.12$, 95\% CI: $[-0.63, 0.87]$; underinclusion, $\beta = 0.43$, 95\% CI: $[-0.58, 1.41]$) and the spirit direction (compliance, $\beta = 0.06$, 95\% CI: $[-0.64, 0.81]$; noncompliance, $\beta = 0.08$, 95\% CI: $[-1.43, 1.62]$; overinclusion, $\beta = 0.00089$, 95\% CI: $[-0.76, 0.74]$; underinclusion, $\beta = 0.33$, 95\% CI: $[-0.80, 1.53]$). Besides these control questions, we also evaluated the base and adapted models on the Massive Multitask Language Understanding (MMLU) benchmark \cite{hendrycks2021measuring}, and did not find degradation on general commonsense knowledge (Table \ref{tab:mmlu-llama}).

\begin{figure}
    \centering
    \includegraphics[width=\linewidth]{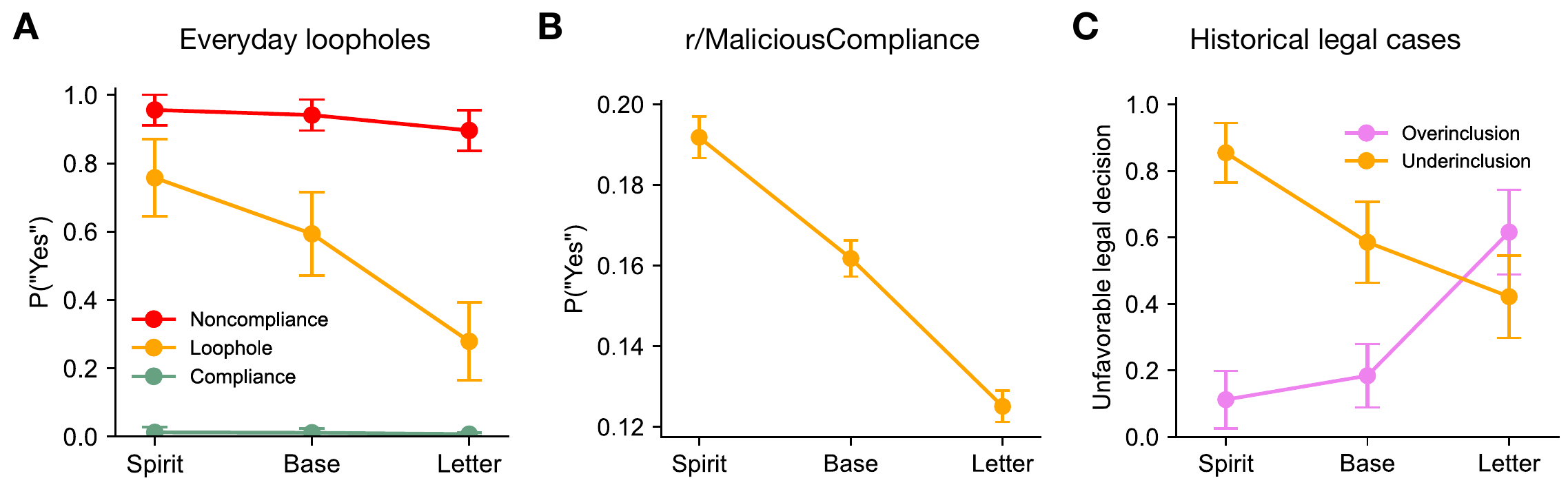}
    \caption{\textbf{Generalization performance of trained letter/spirit adapters (applied at Layer 26) for \texttt{Llama-3.1-70B-Instruct}.} The three panels show targeted directional effect of the adapters on (\textbf{A}) Loophole dataset, (\textbf{B}) Reddit r/MaliciousCompliance dataset, and (\textbf{C}) Historical legal cases. The layout of the figure is similar to Fig. \ref{fig:rs-across-contexts}. Error bars represent 95\% Confidence Intervals. }
    \label{fig:llama-eval-rs-panel}
\end{figure}

In addition, we evaluated generalizations of the letter-spirit adapters for Llama-3.1-70B-Instruct on the Loophole stimuli set. As shown in Fig. \ref{fig:llama-eval-rs-panel}A, the adapted model judged a loophole behavior as less of a disobedience when the letter-oriented adapter was applied ($\beta = -1.14$, 95\% CI: $[-1.43, -0.85]$), and more so when the spirit-oriented adapter was applied ($\beta=0.58$, 95\% CI: $[0.29, 0.87]$). For compliance behavior, the adapters had no significant effect compared to the judgment of base model ($\beta_\text{letter} = 0.048$, 95\% CI: $[-0.23,0.34]$; $\beta_\text{spirit} = -0.047$, 95\% CI: $[-0.33,0.24]$). For noncompliance, we observed a significant decrease of disobedience judgment probability under letter-oriented adapter ($\beta = -0.66$, 95\% CI: $[-1.06, -0.27]$) and a significant increase of judgment probability under spirit-oriented adapter ($\beta = 0.53$, 95\% CI: $[0.01, 1.05]$), although the numerical differences on the response scale are small. When directly comparing the judgment difference between noncompliance and loopholes, we found that the difference is significantly larger in the letter direction than in the spirit direction ($\delta = 0.31$ on the response scale, 95\% CI: $[0.24, 0.38]$).

On the Reddit post dataset from the subreddit r/MaliciousCompliance, we found significant effects of the trained adapters. Compared to the base model, letter-oriented adapter steered the model to judge malicious compliance less as disobedience ($\beta = -0.44$, 95\% CI: $[-0.46,-0.42]$), while spirit-oriented adapter steered the model to judge malicious compliance more as disobedience ($\beta = 0.23$, 95\% CI: $[0.21,0.25]$). The results are visualized in Fig. \ref{fig:llama-eval-rs-panel}B.

We further evaluated the trained letter-spirit adapters for Llama-3.1-70B-Instruct on the Historical Legal Case dataset. We elicited free-form responses from the base and adapted models. The sampled responses were then annotated by an off-the-shelf Qwen3.5-27B model instance. The results were shown in Fig. \ref{fig:llama-eval-rs-panel}C. Compared to base model judgment, overinclusion cases were more likely to receive unfavorable rulings when the letter adapter was applied ($\beta = 3.45$, 95\% CI: $[2.79, 4.16]$). The effect was numerically in the opposite direction when the spirit adapter was applied, although the effect was not significant ($\beta = -0.59$, 95\% CI: $[-1.28, 0.07]$) as the base model had already displayed a strong preference to rule in favor of overinclusion compared to underinclusion ($\beta_\text{base, underinclusion} = 4.97$, 95\% CI: $[3.14, 6.96]$, with overinclusion as the reference level). In contrast, underinclusion cases were significantly less likely to be ruled as a violation of the statute or receive unfavorable judgment, when the letter-oriented adapter was applied ($\beta = -2.41$, 95\% CI: $[-3.11, -1.75]$), and more so when the spirit-oriented adapter was applied ($\beta =  2.09$, 95\% CI: $[1.41, 2.85]$).

\subsubsection*{Model Behaviors on the Control Questions of Rule Vignettes}

\begin{figure}
    \centering
    \includegraphics[width=\linewidth]{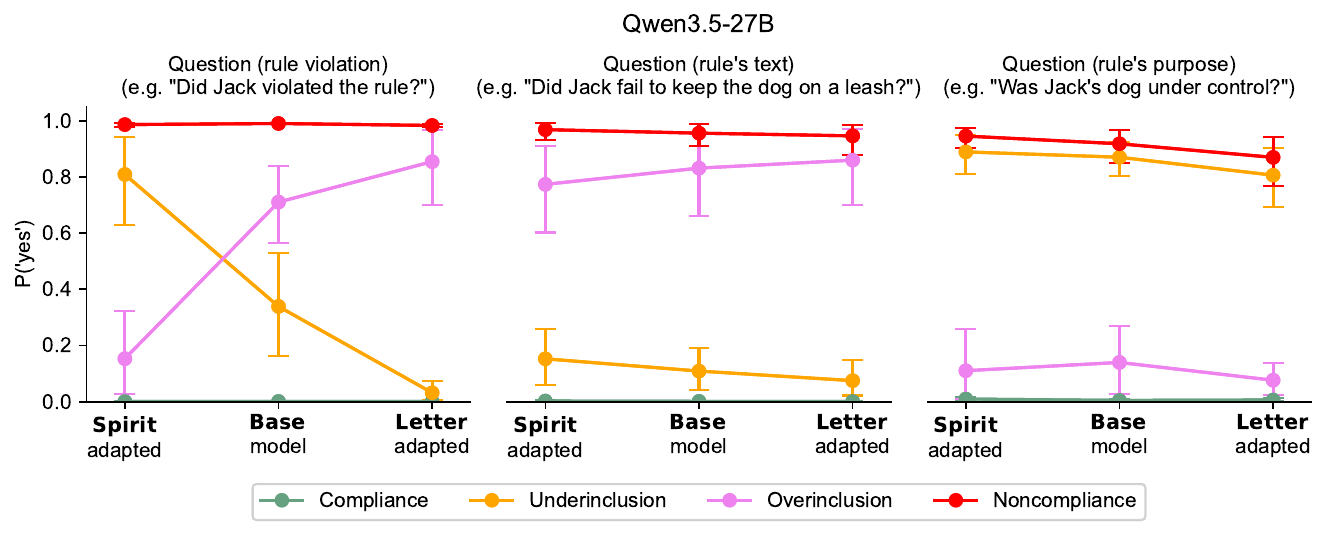}
    \caption{\textbf{Comparing adapted models across different types of questions on held-out data for \texttt{Qwen3.5-27B}.} Base models and adapted models were evaluated on the main question (rule violation, left panel) and two controlled questions (rule's text and rule's purpose, middle and right panel). The structure of the figure is similar to Fig. \ref{fig:rule-question-comparison-llama}.  Error bars represent 95\% Confidence Intervals.}
    \label{fig:rule-question-comparison-qwen}
\end{figure}

Fig. \ref{fig:rule-question-comparison-qwen} compares the effect of \modelname{} adapters on Qwen3.5-27B across the target question (rule violation judgment) and control questions (understanding of a rule's text or purpose). When prompted to judge aspects of a scenario related to the rule's text (Fig. \ref{fig:rule-question-comparison-qwen}B), the adapted model responded similarly to the base model regardless of whether the letter adapter was applied (compliance, $\beta = -0.09$, 95\% CI: $[-0.81, 0.62]$; noncompliance, $\beta = -0.05$, 95\% CI: $[-0.79,0.66]$; overinclusion, $\beta = 0.21$, 95\% CI: $[-0.41,    0.84]$; underinclusion, $\beta = -0.22$, 95\% CI: $[-0.95, 0.47]$) or the spirit adapter was applied (compliance, $\beta = 0.10$, 95\% CI: $[-0.63, 0.82]$; noncompliance, $\beta = 0.23$, 95\% CI: $[-0.53, 0.94]$; overinclusion, $\beta = -0.35$, 95\% CI: $[-0.95, 0.25]$; underinclusion, $\beta = 0.20$, 95\% CI: $[-0.55, 0.90]$). For questions related to the rule's purpose (Fig. \ref{fig:rule-question-comparison-qwen}C), we also did not find any significant effect of the adapters in either the letter (compliance, $\beta = -0.02$, 95\% CI: $[-0.75, 0.70]$; noncompliance, $\beta = -0.28$, 95\% CI: $[-0.94, 0.41]$; overinclusion, $\beta = -0.13$, 95\% CI: $[-0.86, 0.57]$; underinclusion, $\beta = -0.21$, 95\% CI: $[-0.85, 0.45]$) or spirit direction (compliance, $\beta = 0.14$, 95\% CI: $[-0.54, 0.85]$; noncompliance, $\beta = 0.16$, 95\% CI: $[-0.54, 0.85]$; overinclusion, $\beta = -0.18$, 95\% CI: $[-0.88, 0.53]$; underinclusion, $\beta = 0.21$, 95\% CI: $[-0.47,0.87]$). These results suggest that the observed effect on rule violation judgment is targeted without percolating into other topically relevant tasks.

\subsubsection*{Effect of the Adapters on Thinking-enabled \texttt{Qwen3.5-27B}}

\begin{figure}
    \centering
    \includegraphics[width=0.45\linewidth]{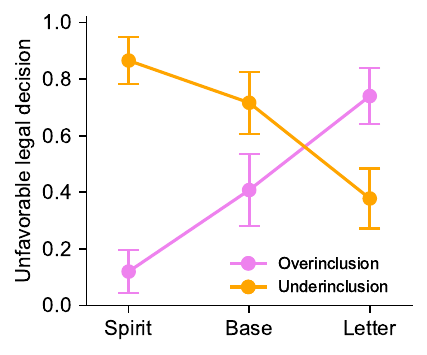}
    \caption{\textbf{Effect of letter/spirit adapters (Layer 32) on free-form legal judgment for historical legal cases with reasoning-enabled \texttt{Qwen3.5-27B}.}  The $y$-axis represents the proportion of unfavorable legal decisions for the specific party in the case. Error bars represent 95\% Confidence Intervals.}
    \label{fig:qwen-thinking-caselaw-rs}
\end{figure}

We additionally evaluated letter/spirit \modelname{} adapters on Qwen3.5-27B for the Historical Legal Case stimuli, with the model's native reasoning capacity enabled. As shown in Figure \ref{fig:qwen-thinking-caselaw-rs}, the letter-oriented adapter steered the model to make fewer unfavorable rulings against underinclusion cases ($\beta=-3.26$, 95\% CI: $[-3.97, -2.61]$) and more unfavorable rulings against overinclusion cases ($\beta=2.77$, 95\% CI: $[2.18, 3.42]$). The pattern was opposite in the spirit direction. With the spirit adapter, the adapted model was more likely to make unfavorable rulings against underinclusion cases ($\beta=2.29$, 95\% CI: $[1.54,3.12]$), and less likely to rule against overinclusion cases ($\beta=-3.38$, 95\% CI: $[    -4.25,-2.60]$).

\subsubsection*{Layer-wise Variation in the Generalization of Trained Adapters}

\begin{figure}
    \centering
    \includegraphics[width=\linewidth]{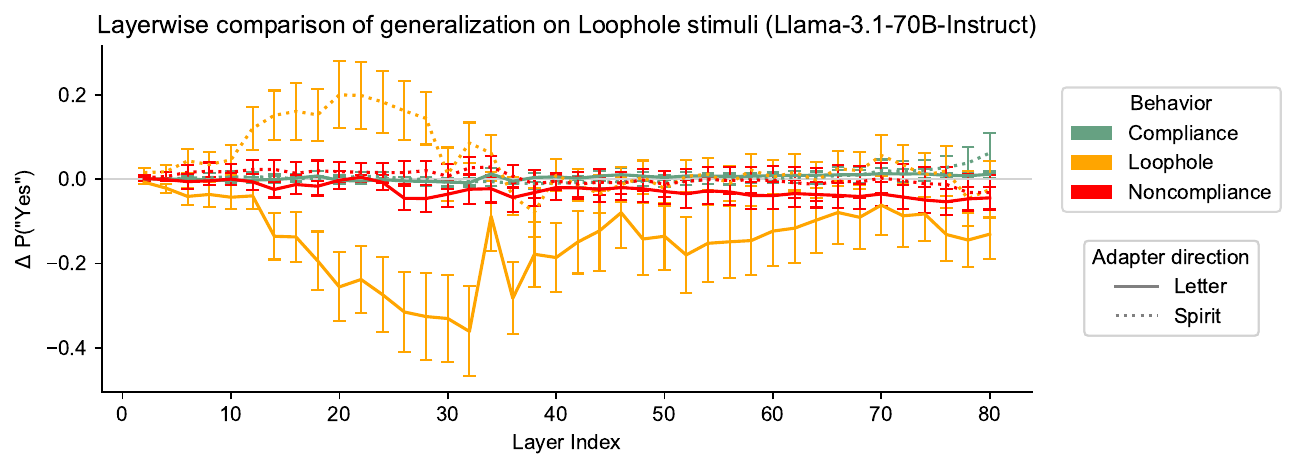}
    \caption{\textbf{Comparing the effect of trained letter/spirit adapters on out-of-distribution Loophole stimuli for \texttt{Llama-3.1-70B-Instruct}.} The $y$-axis represents the changes in the model's likelihood of judging a specific type of behaviors as disobedience (i.e. the probability difference in predicting the token `Yes' as the response to the question ``Did [actor] disobey the request of [actor]'s [other\_person]?'', after applying a trained layer-specific adapter in the direction of letter or spirit). Solid lines show the results of the letter direction, and dotted lines show the results of the spirit direction. Error bars represent 95\% Confidence Intervals.}
    \label{fig:loophole-delta-effect-across-layers-llama}
\end{figure}

\begin{figure}
    \centering
    \includegraphics[width=\linewidth]{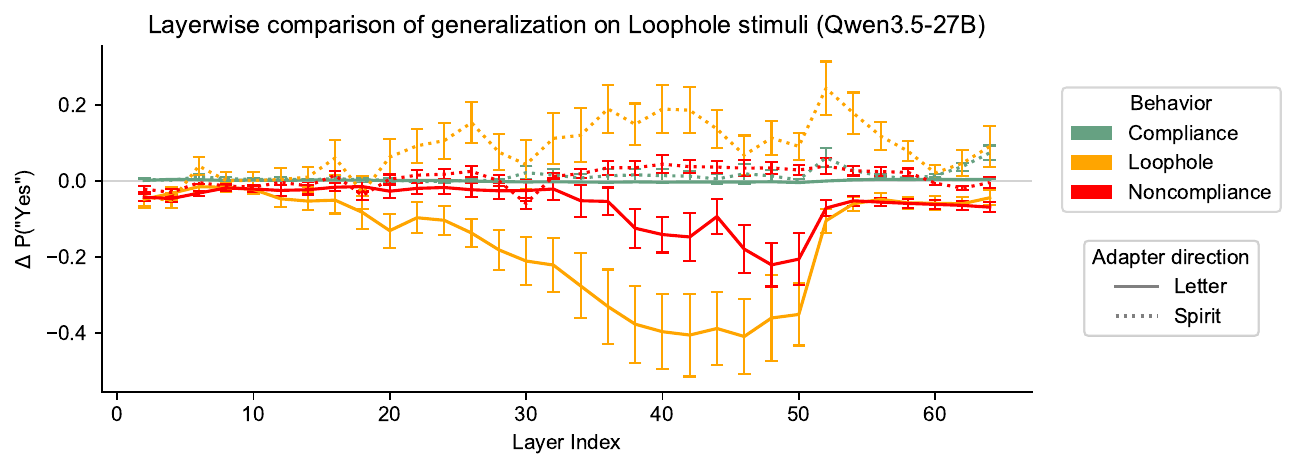}
    \caption{\textbf{Comparing the effect of trained adapters on out-of-distribution Loophole stimuli for \texttt{Qwen3.5-27B}.} The structure of the figure is similar to Fig. \ref{fig:loophole-delta-effect-across-layers-llama}. Error bars represent 95\% Confidence Intervals.}
    \label{fig:loophole-delta-effect-across-layers-qwen}
\end{figure}

We trained layer-specific \modelname{} adapters across different layers of the base model, in order to examine whether the emerging representation of the letter and spirit of the law would be localized in a specific processing stage of the model. For both Qwen3.5-27B and Llama-3.1-70B-Instruct, we trained \modelname{} adapters in the letter and spirit direction separately at every other layer, following the same training protocol (see Materials and Methods for details). All adapters were parameterized with the same rank of 16. Fig. \ref{fig:loophole-delta-effect-across-layers-llama} and \ref{fig:loophole-delta-effect-across-layers-qwen} show the generalization results on the Loophole stimuli. Since the adapters were trained on the Rule dataset, the Loophole dataset, as a set of statistically out-of-distribution stimuli, provides a stronger test of the adapter's generalization. Across the layers of Llama-3.1-70B-Instruct, strong generalization effect occurred around the early middle layers (Layers 14--28). Across the layers of Qwen3.5-27B, robust and targeted generalization effect occurred around the middle layers (Layers 24--36). Based on these results, we identified Layer 26 for Llama-3.1-70B-Instruct and Layer 32 for Qwen3.5-27B as the target layer for adapter intervention, and focused on model behaviors with the letter or spirit adapter activated at the corresponding layer respectively.

\subsubsection*{Inference-time Scaling of Adapter Weights}

We scaled the trained adapter weights during inference time, with varying intensity and sign, as shown in Fig. \ref{fig:steering-comparison-held-out-rule-all} and \ref{fig:steering-comparison-loophole-all}. In addition to the graded steering effect, we also observed that subtracting an adapter from the base model can produce an effect qualitatively similar to adding a trained adapter of the opposite direction. For example, consider the spirit-oriented adapter for Qwen3.5-27B on the held-out rule stimuli (bottom row of Fig. \ref{fig:steering-comparison-held-out-rule-all}). When the weights of the spirit-oriented adapter were 100\% scaled and added to the corresponding layer, the likelihood of being judged as rule violation decreased for overinclusion and increased for underinclusion. Surprisingly, when the weights of the Spirit LoRA were 100\% scaled but \textit{subtracted} from the exact same layer, the adapted model behaved as if it was steered into the letter direction, with the likelihood of being judged as rule violation increased for overinclusion and significantly decreased for underinclusion. Comparing the subtraction and addition of the same adapter, we found significant effects for overinclusion ($\beta = 3.19$, 95\% CI: $[2.00, 4.35]$) and underinclusion ($\beta = -2.43$, 95\% CI: $[-3.60, -1.25]$), and no significant effect for compliance ($\beta = -0.37$, 95\% CI: $[-1.12, 0.41]$) and noncompliance ($\beta = 0.89$, 95\% CI: $[-0.22,2.00]$). Similar effects were also observed when comparing the subtraction versus addition of a 100\% scaled letter-oriented adapter at the same layer (compliance, $\beta = 0.08$, 95\% CI: $[-0.64,0.81]$; noncompliance, $\beta = 0.19$, 95\% CI: $[-0.89,1.21]$; overinclusion, $\beta = -1.47$, 95\% CI: $[-2.58, -0.36]$; underinclusion, $\beta = 2.91$, 95\% CI: $[1.78, 4.05]$).

\subsection*{Contrastive Activation Addition (CAA) Steering}

We conducted Contrastive Activation Addition (CAA) steering \cite{rimsky2024steering} on Llama-3.1-70B-instruct and Qwen3.5-27B, and compared the steering results to that of the \modelname{} adapters. Grounded on the Linear Representation Hypothesis \cite{park2024linear}, Contrastive Activation Addition, or sometimes referred to as the ``difference-of-means'', has been a popular approach to steering Large Language Models to achieve a targeted behavior. We followed the standard recipe of estimating the steering vector as specified in earlier work \cite{rimsky2024steering}. We constructed the contrastive pairs by appending the answer token `Yes' or `No' as the assistant's response under the letter-oriented judgment versus the spirit-oriented judgment. Note that the expected answer would only differ for overinclusion (break the letter but not the spirit) and underinclusion (maintain the letter but subvert the spirit) under the two judgment approaches in contrast, but not for compliance and noncompliance. As a result, the contrastive pairs that effectively contributed to the estimation of steering direction only included overinclusion and underinclusion scenarios. We extracted layer-specific residual stream activation difference vectors across all the contrastive pairs at the position of the last token, and averaged the activation difference vectors to get the estimated steering vector.

\begin{figure}
	\centering
	\includegraphics[width=\linewidth]{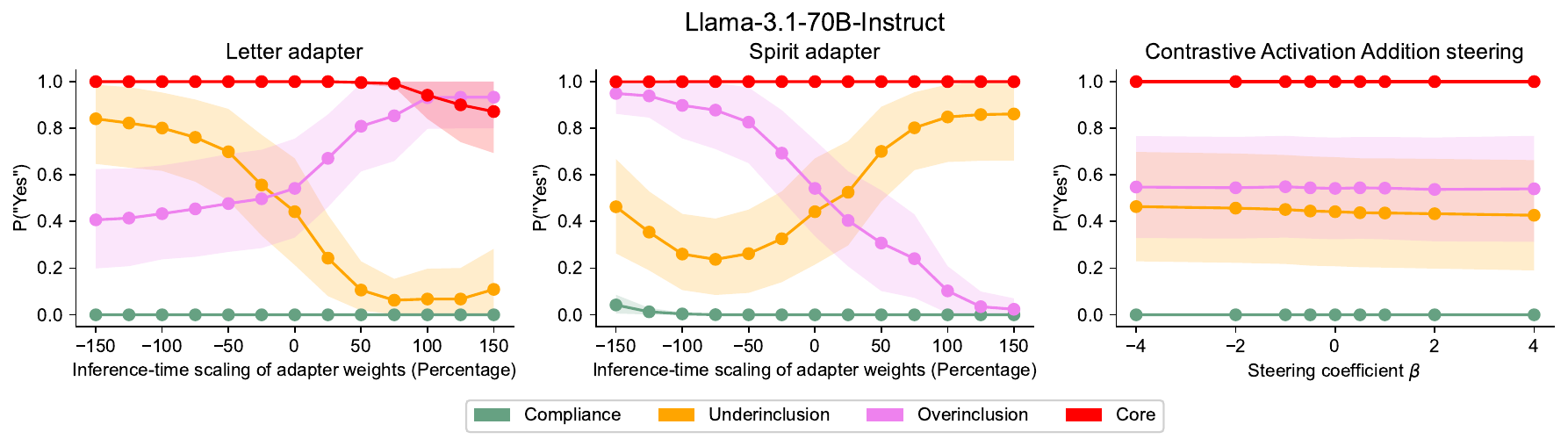}
    \includegraphics[width=\linewidth]{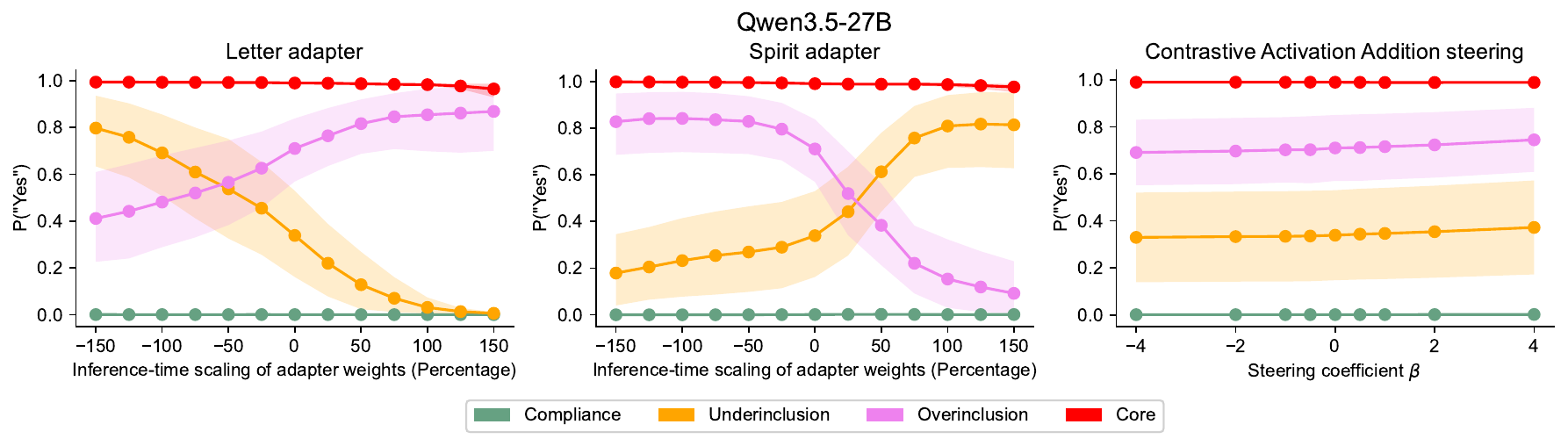}
	\caption{\textbf{Comparing steering effect between scaled \modelname{} adapters and Contrastive Activation Addition on the held-out scenarios from Rule dataset.} The $y$-axis plots the model's likelihood of judging the behavior as rule violation. The plots on the left and middle columns show the steering effect of the letter-oriented adapter and spirit-oriented adapter with different inference-time scaling of the trained adapter's weights. The plots on the right column show the effect of Contrastive Activation Addition approach, based on the same set of contrastive pairs used for training the adapters. The top row shows results on \texttt{Llama-3.1-70B-Instruct}, and the bottom row shows results on \texttt{Qwen3.5-27B}.
        }
	\label{fig:steering-comparison-held-out-rule-all}
\end{figure}

\begin{figure}
	\centering
    \textsf{\footnotesize Llama-3.1-70B-Instruct}\\
        
	\includegraphics[width=\linewidth]{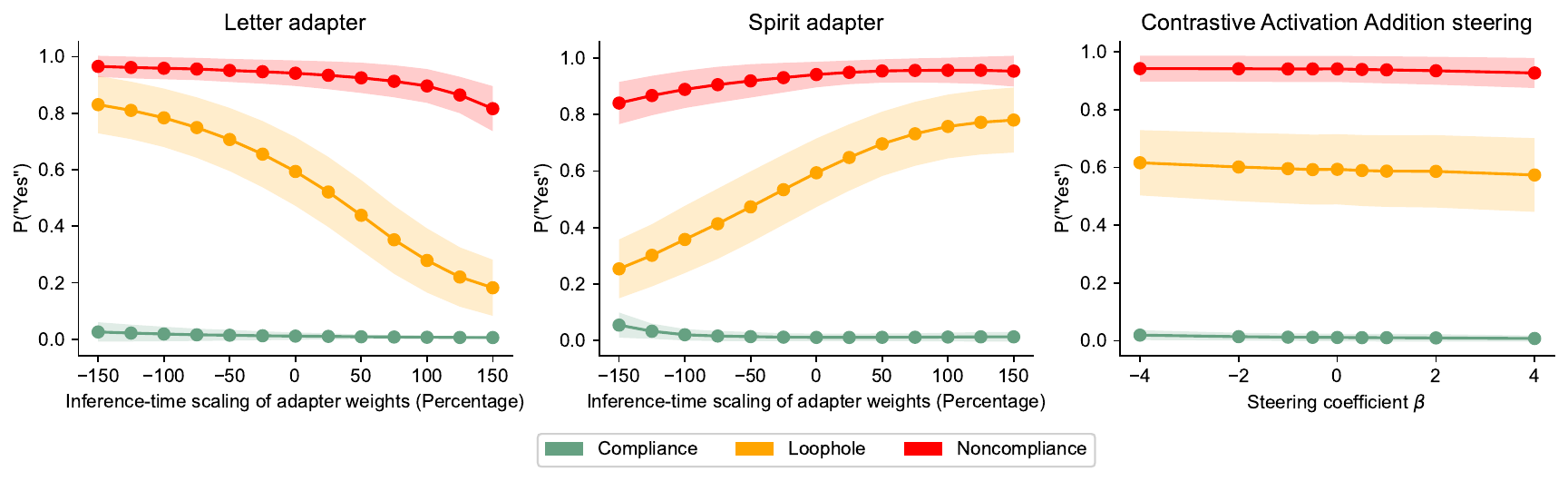}
    
    \textsf{\footnotesize Qwen3.5-27B}\\
    
    \includegraphics[width=\linewidth]{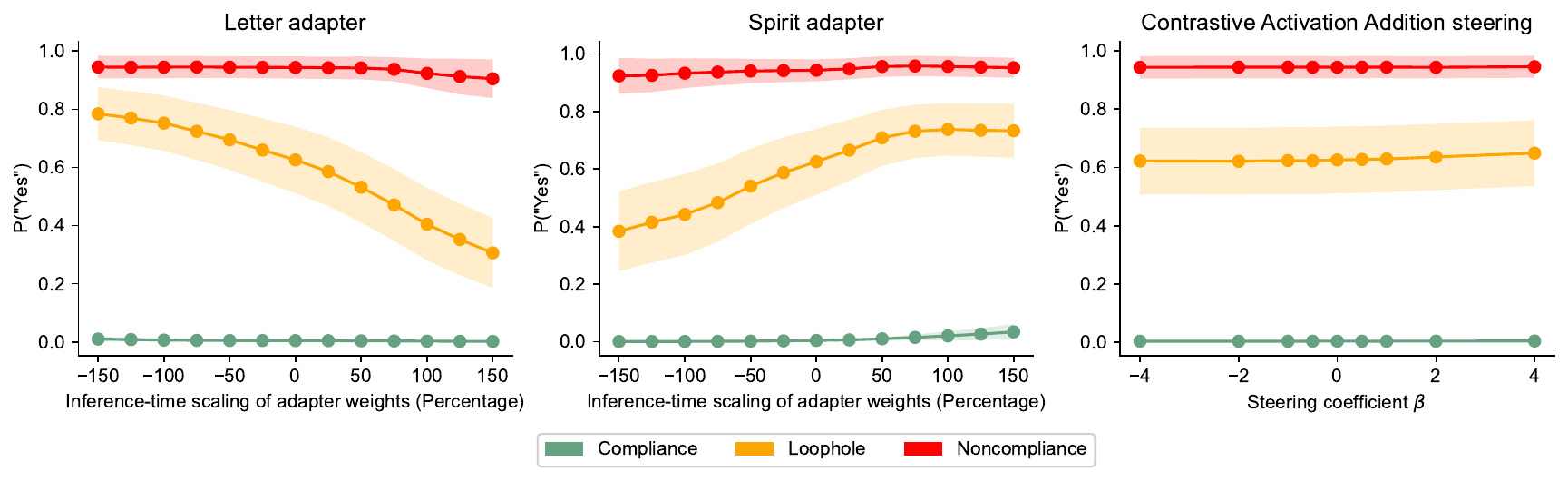}
	\caption{\textbf{Comparing steering effect between scaled \modelname{} adapters and Contrastive Activation Addition on the Loophole vignette dataset.} The left and middle panels show the steering effect of the letter-oriented adapter and spirit-oriented adapter with different inference-time scaling of the trained adapter's weights. The top row shows results on \texttt{Llama-3.1-70B-Instruct}, and the bottom row shows results on \texttt{Qwen3.5-27B}.
    }
	\label{fig:steering-comparison-loophole-all}
\end{figure}

\begin{figure}
    \centering
    \textsf{\footnotesize Llama-3.1-70B-Instruct}\\
    \includegraphics[width=0.48\linewidth]{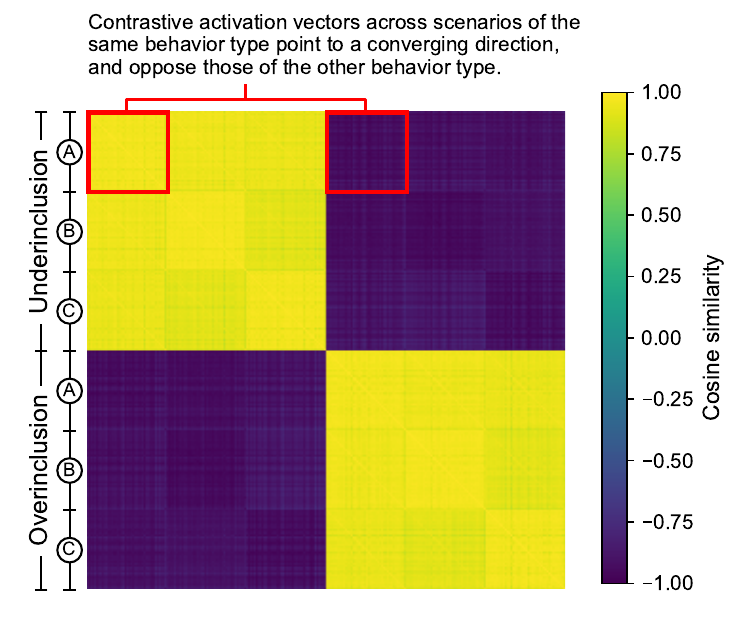}
    \includegraphics[width=0.42\linewidth]{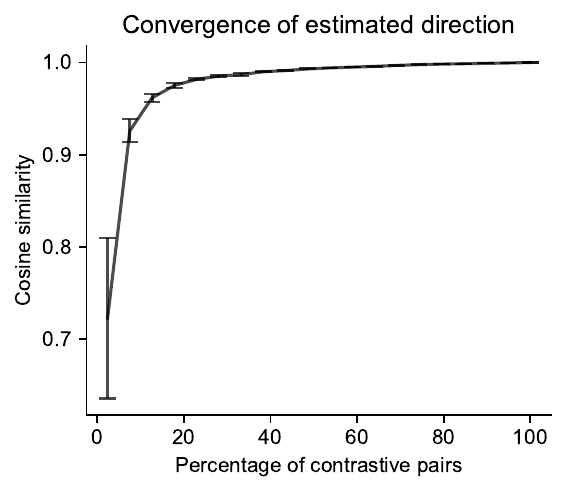}
    
    \textsf{\footnotesize Qwen3.5-27B}\\
    \includegraphics[width=0.48\linewidth]{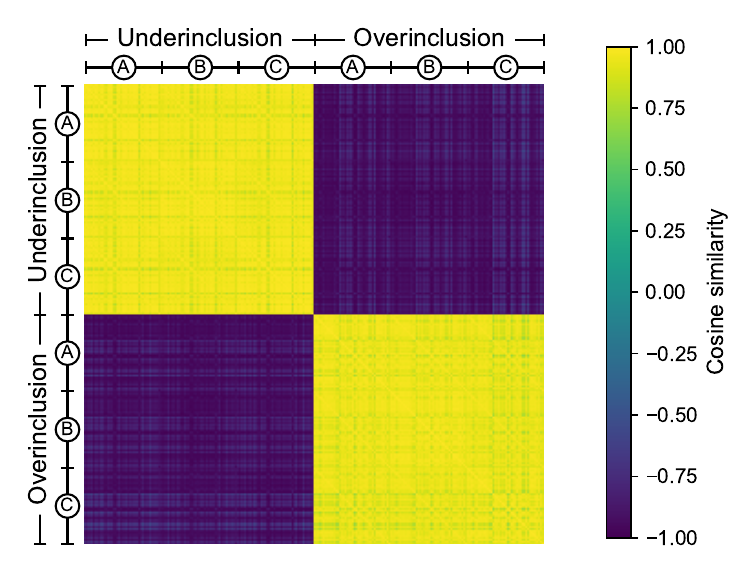}
    \includegraphics[width=0.42\linewidth]{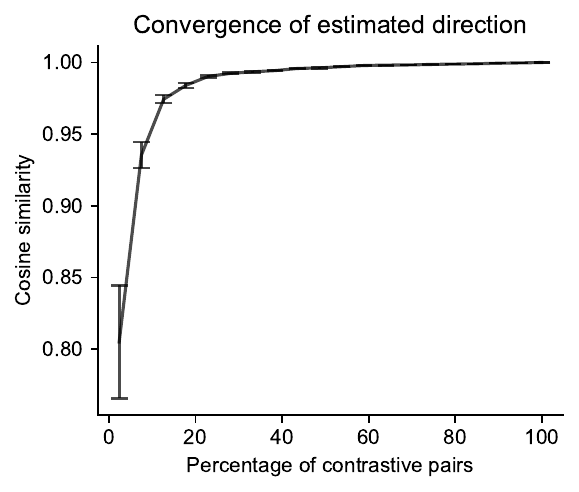}
    \caption{\textbf{Comparing contrastive activation difference vectors within and between different types of behaviors.} We computed the contrastive activation vectors for each contrastive pair in the training set based on 42 rule scenarios and 3 prompt variations, and calculated the cosine similarity among the activation difference vectors. The results (left column) reveal a salient grouping structure predicted by a theoretical model of the conceptual space illustrated in Figure \ref{fig:representational-geometry-rs-panel}. The top row shows the analysis results based on contrastive activation vectors computed at Layer 26 of \texttt{Llama-3.1-70B-Instruct}. The bottom row shows the analysis results based on contrastive activation vectors computed at Layer 32 of \texttt{Qwen3.5-27B}.}
    \label{fig:contrastive-steering-vec-cosine-sim}
\end{figure}

We did not find a successful steering effect based on the Contrastive Activation Addition approach. As shown in Fig. \ref{fig:steering-comparison-held-out-rule-all} and \ref{fig:steering-comparison-loophole-all}, the steered models respond to the critical cases (underinclusion and overinclusion) in a similar way to the base model across steering coefficients of different signs and sizes. There was no interaction between behavior types and steering coefficients. We further analyzed the geometric structure of the steering vectors estimated from the contrastive pairs, as shown in Fig. \ref{fig:contrastive-steering-vec-cosine-sim}. Consistent with a formal model of the relevant dimensions in the conceptual space of jurisprudence approaches, we found that the contrastive activation vectors pointed to opposite directions for overinclusion and underinclusion, as measured by cosine similarity. This observation cannot be attributed to the lack of sufficient contrastive pairs, as suggested by the convergence curve of the contrastive activation vectors (Fig. \ref{fig:contrastive-steering-vec-cosine-sim}).

\newpage

\begin{figure}
    \centering
    \includegraphics[width=0.8\linewidth]{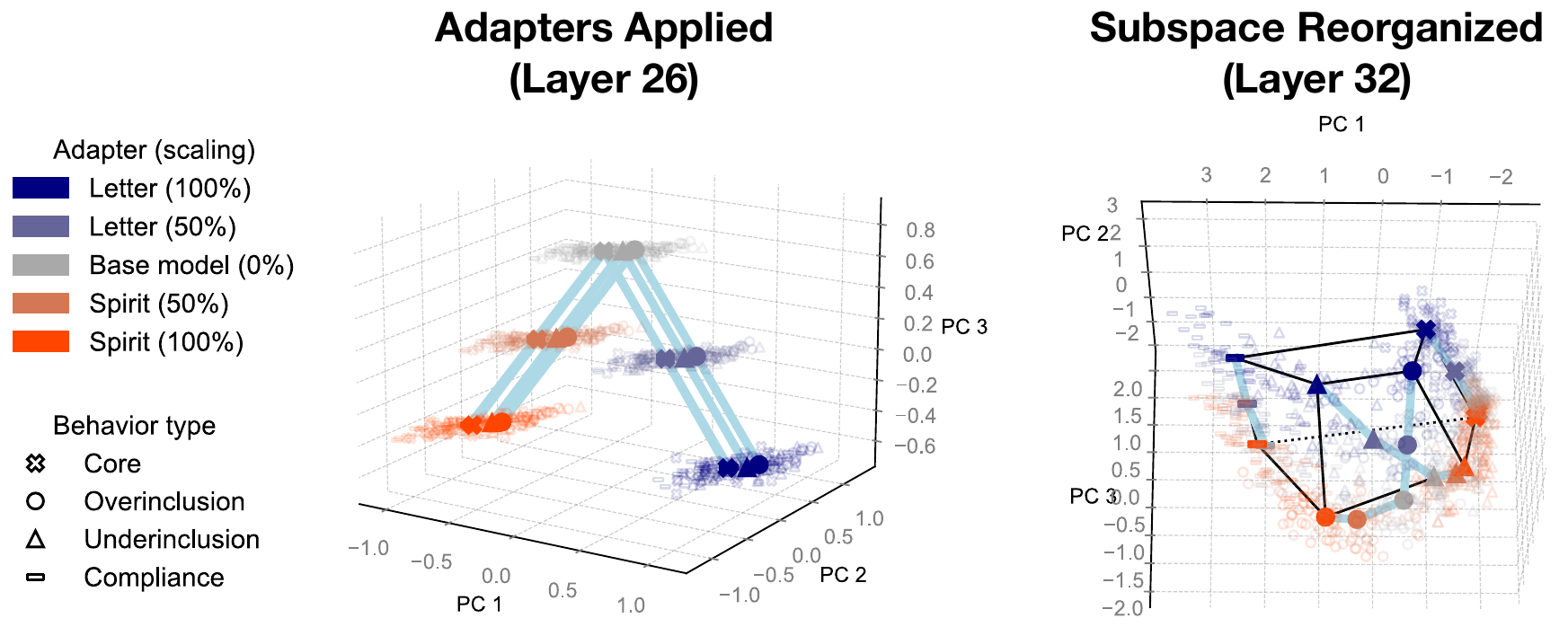}
    \caption{\textbf{Low-dimensional visualization of the emerging representational space in \texttt{Llama-3.1-70B-Instruct}.} We performed Principal Component Analysis on the aggregated set of residual stream activation vectors in the base model \texttt{Llama-3.1-70B-Instruct}, as well as model with trained adapters of various amounts of inference-time scaling (50\%, 100\%).}
    \label{fig:hidden-states-visualization-3d-llama}
\end{figure}

\begin{figure}
\textbf{\textsf{\small A}}

    {\centering
    \includegraphics[width=\linewidth]{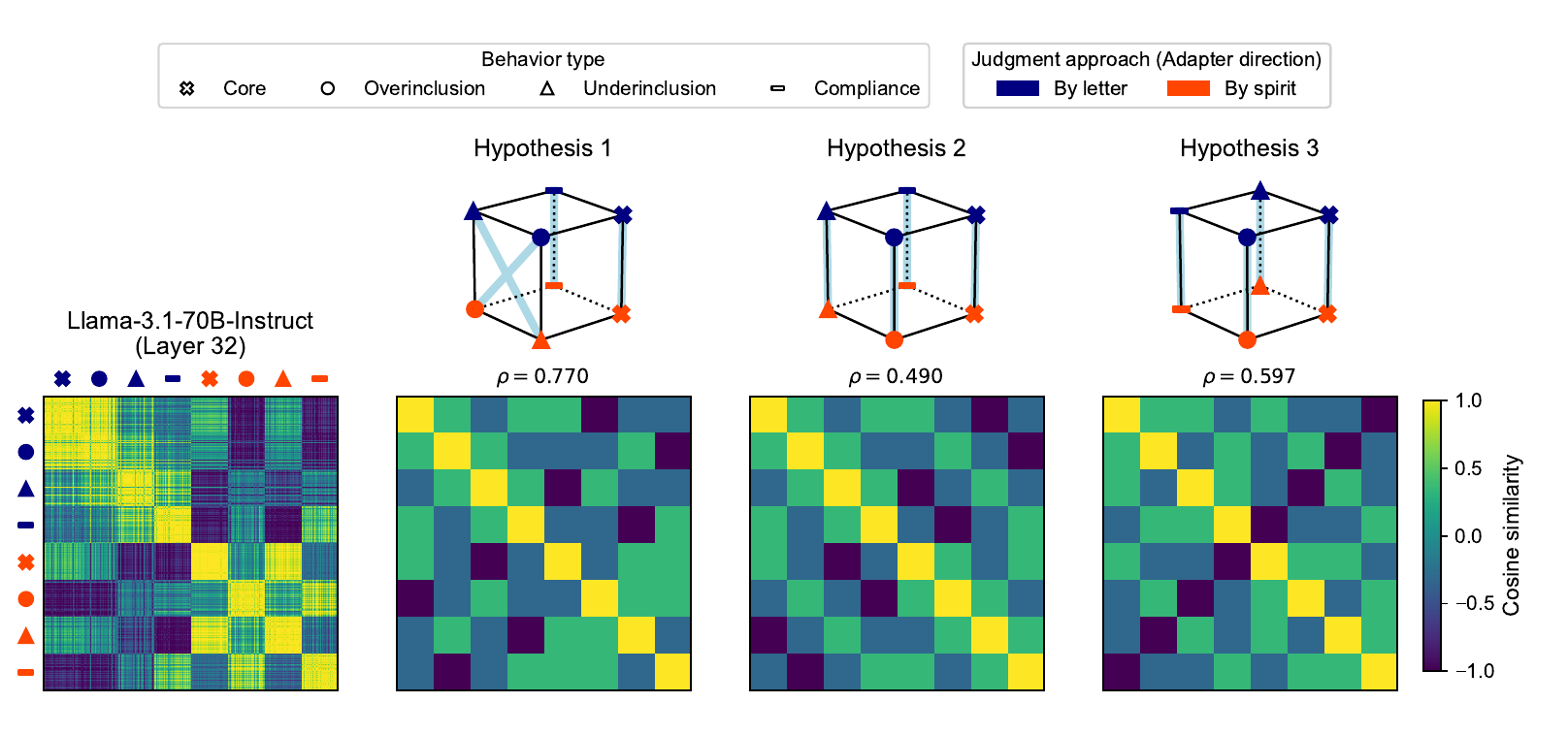}}

\textbf{\textsf{\small B}}

    {\centering
    
    \includegraphics[width=\linewidth]{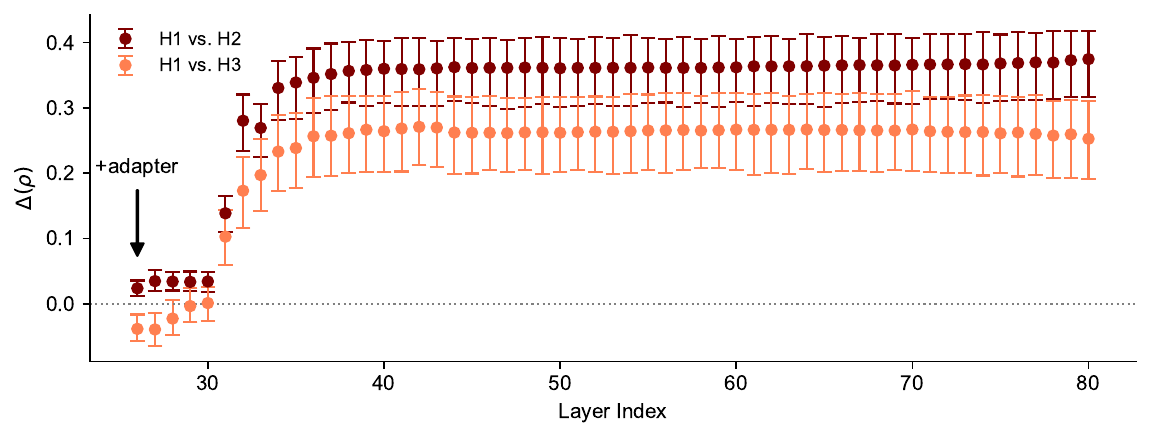}
    \caption{\textbf{Representational Similarity Analysis (RSA) between the hypothesized geometry and empirically-derived low-dimensional subspace from the post-adaptation residual streams of \texttt{Llama-3.1-70B-Instruct}.} \textbf{(A)} Visualization of different hypotheses about possible organization of the representational space for letter-based and spirit-based judgments. We reported Spearman $\rho$ between the Representational Similarity Matrix (RSM) from the first three principal components of model activation vectors to different hypothetical RSMs. \textbf{(B)} Correlation differences between different hypothetical geometry with bootstrapped 95\% Confidence Intervals.}
    \label{fig:rsa-comparison-llama31}
    }
\end{figure}

\begin{figure}
\textbf{\textsf{\small A}}

    {\centering
    \includegraphics[width=\linewidth]{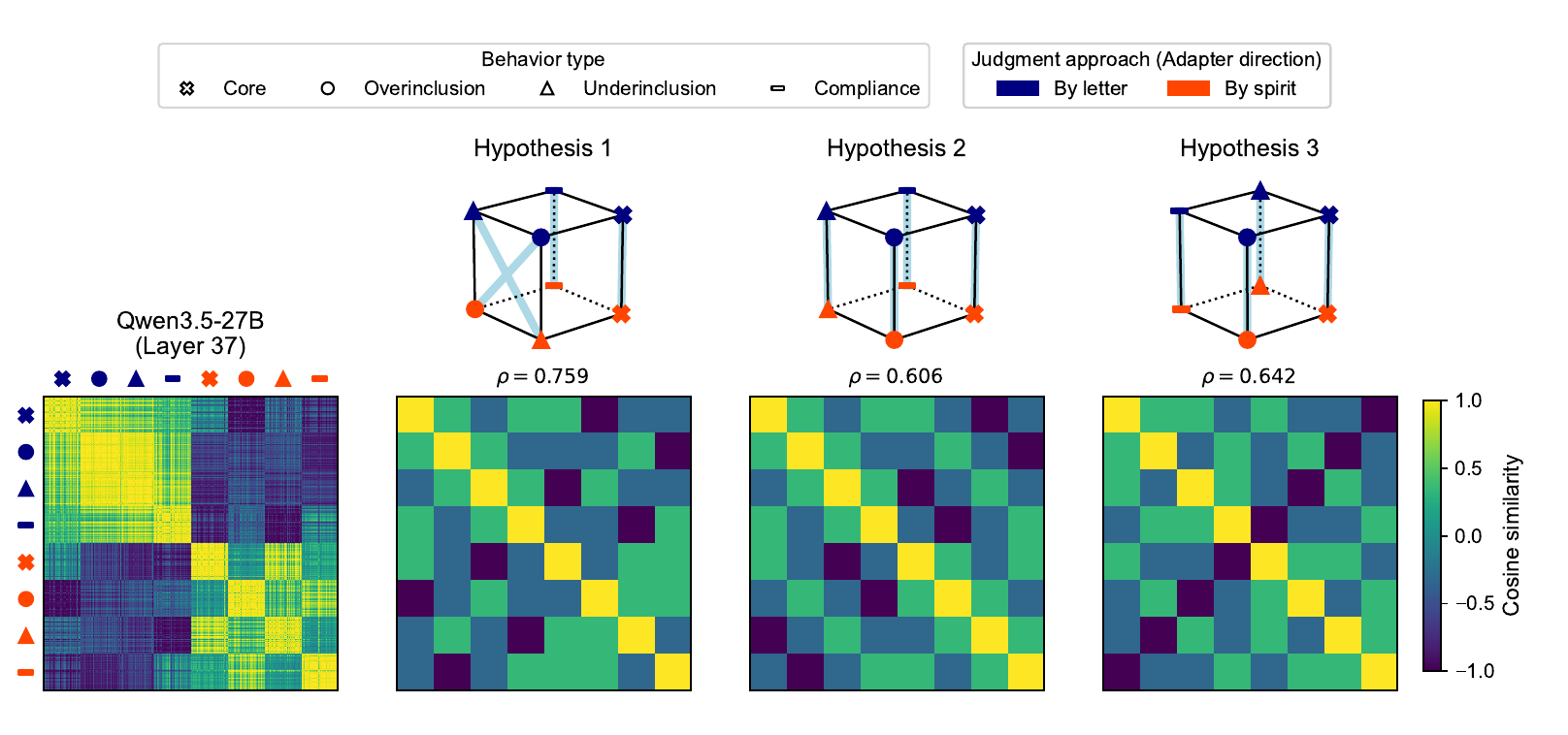}}

\textbf{\textsf{\small B}}

    {\centering
    
    \includegraphics[width=\linewidth]{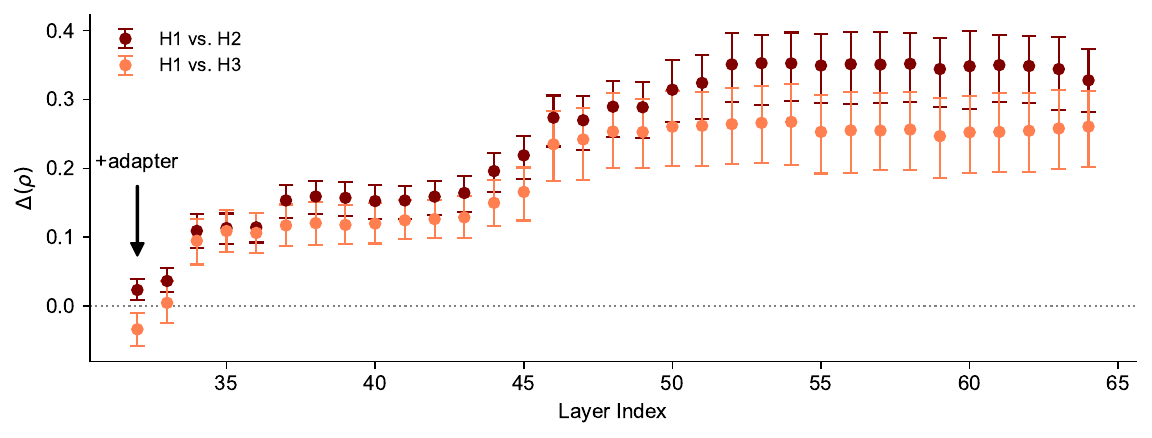}
    \caption{\textbf{Representational Similarity Analysis (RSA) between the hypothesized geometry and empirically derived low-dimensional subspace from the post-adaptation residual streams of \texttt{Qwen3.5-27B}.} \textbf{(A)} Visualization of different hypotheses about possible organization of the representational space for letter-based and spirit-based judgments. We reported Spearman $\rho$ between the Representational Similarity Matrix (RSM) from the first three principal components of model activation vectors to different hypothetical RSMs. \textbf{(B)} Correlation differences between different hypothetical geometries with bootstrapped 95\% Confidence Intervals.}
    \label{fig:rsa-comparison-qwen35}
    }
\end{figure}

\clearpage

\begin{table}
	\centering
	\caption{\textbf{Accuracies on Massive Multitask Language Understanding (MMLU) benchmark for base model Llama-3.1-70B-Instruct as well as with letter/spirit adapters.} The columns correspond to the specific model (with or without LoRA) being evaluated. Each row reports the accuracy of the model on the Massive Multitask Language Understanding (MMLU) benchmark, as well as performance on specific law-related sections.
    }
	\label{tab:mmlu-llama}
	
	\begin{tabular}{lccc}
		\\
		\hline
		Section & Spirit adapted & Base model & Letter adapted\\
		\hline
		All subjects & 85.61 & 85.59 & 85.56\\
		International law & 90.08 & 91.74 & 90.08\\
		Jurisprudence & 87.96 & 87.03 & 88.89\\
        Professional law & 65.06 & 65.31 & 64.66\\
		\hline
	\end{tabular}
\end{table}

\begin{table}
	\centering

	\caption{\textbf{Accuracies on Massive Multitask Language Understanding (MMLU) benchmark for base model Qwen3.5-27B as well as with letter/spirit adapters.} The structure of the table is the same as Table \ref{tab:mmlu-llama}.}
	\label{tab:mmlu-qwen35}
	
	\begin{tabular}{lccc}
		\\
		\hline
		Section & Spirit adapted & Base model & Letter adapted\\
		\hline
		All subjects & 88.93  &  88.88 & 89.06 \\
		International law &  95.04 & 94.21  & 95.87\\
		Jurisprudence & 91.67  &  88.89 & 89.81\\
        Professional law & 74.32 &  73.86 & 74.45\\
		\hline
	\end{tabular}
\end{table}

\end{document}